%% file: sasvi.tex
\documentclass{article} 

\usepackage{times}
\usepackage{graphicx}
\usepackage{amsmath}
\usepackage{bm}
\usepackage{url}
\usepackage{amssymb}

\usepackage{multirow}

\newcounter{counter_theorem}
\newcounter{counter_lemma}

\newtheorem{theorem}[counter_theorem]{Theorem}
\newtheorem{lemma}[counter_lemma]{Lemma}

\title{Safe Screening with Variational Inequalities and Its Application to Lasso}

\begin{document}

\author{Jun Liu$^1$, Zheng Zhao$^1$, Jie Wang$^2$, and Jieping Ye$^2$ \\ 
        $^1$SAS Institute Inc. \\ 
				$^2$Arizona State University \\
				\{jun.liu,zheng.zhao\}@sas.com \\
				\{jie.wang.ustc,jieping.ye\}@asu.edu
}

\maketitle

\begin{abstract}
Sparse learning techniques have been routinely used for feature selection as the resulting model
usually has a small number of non-zero entries.  
Safe screening, which eliminates the features that are guaranteed to have zero coefficients for 
a certain value of the regularization parameter, is a technique for improving the computational 
efficiency. Safe screening is gaining increasing attention since 1) solving sparse learning formulations
usually has a high computational cost especially when the number of features is large and 
2) one needs to try several regularization parameters to select a 
suitable model.
In this paper, we propose an approach called ``Sasvi" (Safe screening with variational inequalities). 
Sasvi makes use of the variational inequality that provides the sufficient and necessary optimality condition for the dual problem. 
Several existing approaches for Lasso screening can be casted as relaxed versions of the proposed Sasvi, thus Sasvi provides a stronger safe screening rule. 
We further study the monotone properties of Sasvi for Lasso, based on which a sure removal regularization parameter can be identified for each feature. 
Experimental results on both synthetic and real data sets are reported to demonstrate the effectiveness of the proposed Sasvi for Lasso screening.
\end{abstract}

\section{Introduction}

\input{introduction}

\section{The Proposed Sasvi}

\input{lasso}

\section{Comparison with Existing Approaches}\label{s:comparison}

\input{relationship}

\section{Feature Sure Removal Parameter}\label{s:sure:remove}

\input{sureRemove}





\section{Experiment}\label{s:experiment}

\input{experiment2}

\section{Conclusion}

The safe screening is a technique for improving the computational efficiency 
by eliminating the inactive features in sparse learning algorithms. 
In this paper, we propose a novel approach called Sasvi (Safe screening with variational inequalities).
The proposed Sasvi has three modules: 
dual problem derivation, feasible set construction, and upper-bound estimation.
The key contribution of
the proposed Sasvi is the usage of the variational inequality 
which provides the sufficient and necessary optimality conditions for the dual problem.
Several existing approaches can be casted as relaxed versions of the proposed Sasvi, and thus 
Sasvi provides a stronger screening rule. 
The monotone properties of 
the established upper-bound are studied based on a sure removal regularization parameter which can be identified for each feature.

The proposed Sasvi can be extended to solve the generalized sparse linear models, 
by filling in Figure~\ref{fig:sasvi:diagram} with the three key modules.
For example, the sparse logistic regression can be written as
\begin{equation}\label{eq:l1:logistic}
  \min_{\bm \beta}   \sum_{i=1}^n \log (1+ \exp (-y_i   \bm \beta ^T \mathbf x_i  ) ) + \lambda \|\bm \beta\|_1.
\end{equation}
We can derive its dual problem as
\begin{equation*}
	 \min_{\bm \theta: \| X^T \bm \theta \|_{\infty} \le 1 }  - \sum_{i=1}^n \left( \log \left( \frac{\frac{y_i}{\lambda}}{\frac{y_i}{\lambda} - \theta_i} \right)    + \frac{\theta_i }{\frac{y_i}{\lambda}}  \log ( \frac{\frac{y_i}{\lambda} - \theta_i }{ \theta_i} )  \right).
\end{equation*}
According to Lemma~\ref{lemma:optimization}, 
for the dual optimal $\theta_i^*$, 
the optimality condition via the variational inequality is
\begin{equation*} 
	 \sum_{i=1}^n  \frac{1}{ \frac{y_i}{\lambda}}  \log \left( \frac{\frac{y_i}{\lambda} - \theta_i^* }{ \theta_i^*} \right) (\theta_i - \theta_i^*)   \le 0, \forall \bm  \theta: \| X^T \bm \theta \|_{\infty} \le 1.
\end{equation*}
Then, we can construct the feasible set for $\bm \theta_2^*$ at the regularization parameter $\lambda_2$
in a similar way to the $\Omega (\bm \theta_2^*)$ in Eq.~\eqref{eq:feasible:set}. Finally, we can estimate 
the upper-bound of $|\langle  \mathbf x_j ,  \bm \theta^*_2 \rangle| $ by Eq.~\eqref{eq:proposal:II},
and discard the $j$-th feature if such upper-bound is smaller than 1. Note that, 
compared to the Lasso case, Eq.~\eqref{eq:proposal:II} is
much more challenging for the logistic loss case. We plan to replace the feasible 
set $\Omega (\bm \theta_2^*)$ by its quadratic approximation so that Eq.~\eqref{eq:proposal:II}  has an easy solution.
We also plan to apply the proposed Sasvi to solving the Lasso solution path using LARS.

\clearpage

\onecolumn

\appendix

\noindent \textbf{\huge Supplementary Material}

\input{appendix}

\end{document}

%% file: introduction.tex
Sparse learning~\cite{Candes:2008,Tibshirani:Lasso:1996} is an effective technique for analyzing high dimensional data. 
It has been applied successfully in various areas, such as machine learning, signal processing, image processing,
medical imaging, and so on. In general, the $\ell_1$-regularized sparse learning can be formulated as: 
\begin{equation}\label{eq:l1:sparse:learning}
    \min_{\bm \beta} \quad \mbox{loss}(\bm \beta) + \lambda \|\bm \beta\|_1,
\end{equation}
where $\bm \beta \in \mathbb{R}^p$ contains the model coefficients, 
$\mbox{loss}(\bm \beta)$ is a loss function defined on the design matrix $X \in \mathbb{R}^{n \times p}$ 
and the response $\mathbf y \in  \mathbb{R}^n$,
and $\lambda$ is a positive regularization parameter that balances the tradeoff between the loss function and the $\ell_1$ regularization. 
Let $\mathbf x^i \in \mathbb{R}^p$ denote the $i$-th sample that corresponds to the transpose of the $i$-th row of $X$,
and let $\mathbf x_j \in \mathbb{R}^n$ denote the $j$-th feature that corresponds to the $j$-th column of $X$.
We use 
$\mbox{loss}(\bm \beta) = \frac{1}{2} \|X \bm \beta - \mathbf y \|_2^2= \frac{1}{2} \sum_{i=1}^n (y_i - \bm \beta ^T \mathbf x^i)^2$ in Lasso~\cite{Tibshirani:Lasso:1996} and 
$\mbox{loss}(\bm \beta) = \sum_{i=1}^n \log (1+ \exp (-y_i   \bm \beta ^T \mathbf x^i  ) ) $ in sparse logistic regression~\cite{Koh:sparse:logistic:2007}.

Since the optimal $\lambda$ is usually unknown in practical applications, we need to solve formulation~\eqref{eq:l1:sparse:learning} corresponding to
a series of regularization parameter $\lambda_1 > \lambda_2 > \ldots > \lambda_k$, obtain
the solutions $\bm \beta^*_1, \bm \beta^*_2, \ldots, \bm \beta^*_k$,
and then select the solution that is optimal in terms of a pre-specified criterion, e.g., Schwarz Bayesian information criterion~\cite{Schwarz:SBC:1978} and cross-validation.
The well-known LARS approach~\cite{Efron:LAR:2004} can be modified to obtain the full piecewise linear Lasso solution path. 
Other approaches such as interior point~\cite{Koh:sparse:logistic:2007}, 
coordinate descent~\cite{Friedman:coordinate:2010} 
and accelerated gradient descent~\cite{Nesterov:2004}
usually solve formulation~\eqref{eq:l1:sparse:learning}
corresponding to a series of pre-defined parameters. 

The solutions $\bm \beta^*_k, k=1, 2, \ldots, $ are sparse in that
many of their coefficients are zero. Taking advantage of the nature of sparsity, 
the screening techniques have been proposed for accelerating the computation. 
Specifically, given a solution $\bm \beta^*_1$ at the regularization parameter $\lambda_1$, if
we can identify the features that are guaranteed to have zero coefficients in $\bm \beta^*_2$ at the regularization parameter $\lambda_2$,
then the cost for computing $\bm \beta^*_2$ can be saved by excluding those inactive features.
There are two categories of screening techniques: 1) the safe screening techniques~\cite{Ghaoui:2012,Wang:2012:report,Ogawa2013,Xiang:2011}
with which our obtained solution is exactly the same as 
the one obtained by directly solving \eqref{eq:l1:sparse:learning}, and 2)
the heuristic rule such as the strong rules~\cite{Tibshirani:2012} which can eliminate more features but  
might mistakenly discard active features. 

In this paper, we propose an approach called ``Sasvi" (Safe screening with variational inequalities) and 
take Lasso as an example in the analysis. 
Sasvi makes use of the variational inequality 
which provides the sufficient and necessary optimality condition for the dual problem.
Several existing approaches such as SAFE~\cite{Ghaoui:2012} and DPP \cite{Wang:2012:report}
can be casted as relaxed versions of the proposed Sasvi, thus 
Sasvi provides a stronger screening rule. 
The monotone properties of Sasvi for Lasso are studied based on which a sure removal regularization parameter can be identified for each feature.
Empirical results on both synthetic and real data sets demonstrate the effectiveness of the proposed Sasvi for Lasso screening.
Extension of the proposed Sasvi to the generalized sparse linear models such as logistic regression is briefly discussed.

\noindent \textbf{Notations } Throughout this paper, scalars are denoted by italic letters, and vectors by bold face letters. 
Let $\|\cdot\|_1$, $\|\cdot\|_2$, $\|\cdot\|_{\infty}$ denote the $\ell_1$ norm, the Euclidean norm, and the infinity norm, respectively.
Let $\langle \mathbf x, \mathbf y \rangle$ denote the inner product between $\mathbf x$ and $\mathbf y$.

%% file: lasso.tex
Our proposed approach builds upon an analysis on the following simple problem:
\begin{equation}\label{eq:simple:fact}
 \min_{\beta} \left\{- \beta b + | \beta | \right\}.
\end{equation}
We have the following results: \\
1) If $|b| \le 1$, then the minimum of \eqref{eq:simple:fact} is 0;\\
2) If $|b| > 1$, then the minimum of \eqref{eq:simple:fact} is $-\infty$; and\\
3)  If $|b|<1$, then the optimal solution $\beta^* =0$.

The dual problem usually can provide a good insight about the problem to be solved.
Let $\bm \theta$ denote the dual variable of Eq.~\eqref{eq:l1:sparse:learning}. 
In light of Eq.~\eqref{eq:simple:fact}, we can show that $ \beta_j^*$, the $j$-th
component of the optimal solution to Eq.~\eqref{eq:l1:sparse:learning},
optimizes
\begin{equation}\label{eq:sasvi:j:feature}
 \min_{\beta_j} \left\{- \beta_j \langle \mathbf x_j, \bm \theta^* \rangle + | \beta_j | \right\},
\end{equation}
where $\mathbf x_j$ denotes the $j$-th feature
and $\bm \theta^*$ denotes the optimal dual variable of Eq.~\eqref{eq:l1:sparse:learning}.
From the results to Eq.~\eqref{eq:simple:fact},
we need $|\langle  \mathbf x_j ,  \bm \theta^* \rangle| \le 1$
to ensure that Eq.~\eqref{eq:sasvi:j:feature} does not equal to $-\infty$\footnote{This is used
in deriving the last equality of Eq.~\eqref{eq:Lasso:primal:dual}.},
and we have
\begin{equation}\label{eq:simple:result}
  |\langle  \mathbf x_j ,  \bm \theta^* \rangle| < 1    \Rightarrow   \beta_j^*= 0.
\end{equation}
Eq.~\eqref{eq:simple:result} says that, the $j$-th feature can be safely eliminated
in the computation of  $\bm \beta^*$ if $|\langle  \mathbf x_j ,  \bm \theta^* \rangle| < 1$.

\begin{figure}
\begin{center}
\centerline{\includegraphics[width=0.9\columnwidth]{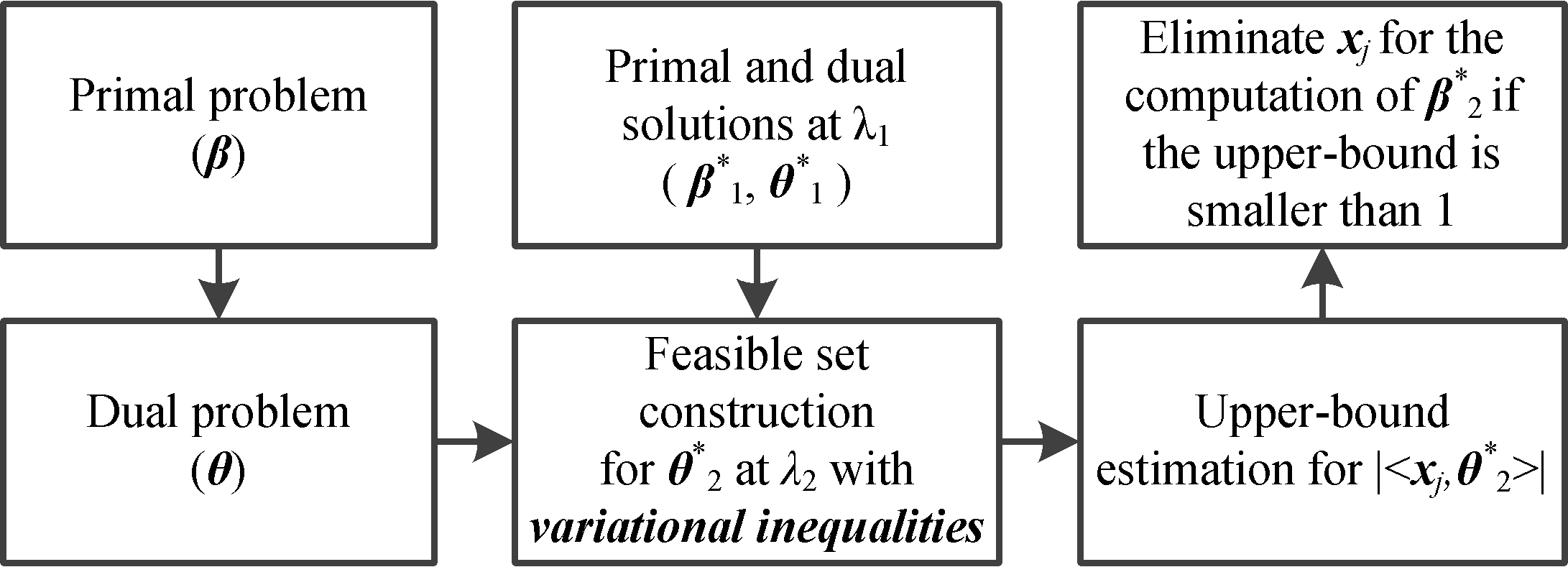}}
\caption{The work flow of the proposed Sasvi. The purpose is to discard the features
that can be safely eliminated in computing $\bm \beta^*_2$ with the information obtained at $\lambda_1$.}\label{fig:sasvi:diagram}
\end{center}
\vskip -0.2in
\end{figure}

Let $\lambda_1$ and $\lambda_2$ be two distinct regularization parameters that satisfy 
\begin{equation}\label{eq:lambda:max}
  \lambda_{\max} \ge \lambda_1 > \lambda_2 >0,
\end{equation}
where $\lambda_{\max}$ denotes the value of $\lambda$ above which the solution to Eq.~\eqref{eq:l1:sparse:learning}
is zero.
Let $\bm \beta_1^*$ and $\bm \beta_2^*$ be the optimal primal variables corresponding to $\lambda_1$ and $\lambda_2$, respectively.
Let $\bm \theta_1^*$ and $\bm \theta_2^*$ be the optimal dual variables corresponding to $\lambda_1$ and $\lambda_2$, respectively. 
Figure~\ref{fig:sasvi:diagram} illustrates the work flow of the proposed Sasvi. 
We firstly derive the dual problem of Eq.~\eqref{eq:l1:sparse:learning}.
Suppose that we have obtained the primal and dual solutions $\bm \beta_1^*$ and $\bm \theta_1^*$ for a given regularization parameter $\lambda_1$,
and we are interested in solving Eq.~\eqref{eq:l1:sparse:learning} with $\lambda=\lambda_2$
by using Eq.~\eqref{eq:simple:result} to screen the features to save computational cost.
However, the difficulty lies in that,  we do not
have the dual optimal $\bm \theta^*_2$.
To deal with this, we construct a feasible set for $\bm \theta^*_2$, estimate an upper-bound of $|\langle  \mathbf x_j ,  \bm \theta^*_2 \rangle|$,
and safely remove $\mathbf x_j$ if this upper-bound is smaller than 1.

The construction of a tight feasible set for $\bm \theta^*_2$ is key to the success of the screening technique. If the constructed 
feasible set is too loose, the estimated upper-bound of $|\langle  \mathbf x_j ,  \bm \theta^*_2 \rangle|$ is over 1, 
and thus only a few features can be discarded.
In this paper, 
we propose to construct the feasible set by using the variational inequalities 
that provide the sufficient and necessary optimality conditions for the dual problems with $\lambda=\lambda_1$ and $\lambda_2$.
Then, we estimate the upper-bound of $|\langle  \mathbf x_j ,  \bm \theta^*_2 \rangle|$ in the constructed feasbile set, and 
discard the $j$-th feature 
if the upper-bound is smaller than 1. 
For discussion convenience, we focus on Lasso in this paper, but the underlying methodology can be extended to the general problem in Eq.~\eqref{eq:l1:sparse:learning}.
Next, we elaborate the three building blocks that are illustrated in the bottom row of Figure~\ref{fig:sasvi:diagram}.

\subsection{The Dual Problem of Lasso}\label{ss:dual:Lasso}

We follow the discussion in Section 6 of~\cite{Nesterov:core:2007}
in deriving the dual problem of Lasso as follows:
\begin{equation}\label{eq:Lasso:primal:dual}
\begin{aligned}
  &     \min_{\bm \beta}                                            \left [\frac{1}{2} \left \|X \bm \beta - \mathbf y \right \|_2^2 + \lambda \|\bm \beta\|_1   \right ]\\
  &  =   \min_{\bm \beta} \max_{\bm \theta}                          \left [ \langle \mathbf y -X \bm \beta, \lambda \bm \theta  \rangle - \frac{1}{2} \| \lambda \bm \theta\|_2^2 + \lambda \|\bm \beta\|_1  \right ] \\
  &  =  \max_{\bm \theta} \min_{\bm \beta}                           \lambda  \left [ \langle \mathbf y, \bm \theta \rangle - \frac{\lambda \| \bm \theta \|_2^2}{2}  - \langle  X^T   \bm \theta, \bm \beta \rangle +   \|\bm \beta\|_1   \right] \\
  &  =  \max_{\bm \theta: \| X^T \bm \theta\|_{\infty} \le 1 }     \lambda^2 \left [ - \frac{1}{2} \left \| \bm \theta- \frac{\mathbf y}{\lambda} \right \|_2^2 + \frac{1}{2} \left \| \frac{\mathbf y}{\lambda} \right \|_2^2 \right ].
\end{aligned}
\end{equation}
A dual variable $\bm \theta$ is introduced in the first equality,
and the equivalence can be verified by setting the derivative with regard to $\bm \theta$ to zero, which leads to the following
relationship between the optimal primal variable ($\bm \beta^*$) and the optimal dual variable ($\bm \theta^*$):
\begin{equation}\label{eq:primal:dual:relationship}
   \lambda \bm \theta^* =  \mathbf y -X \bm \beta^*.
\end{equation}
In obtaining the last equality of Eq.~\eqref{eq:Lasso:primal:dual}, we make use of the 
results to Eq.~\eqref{eq:simple:fact}.

The dual problem of Eq.~\eqref{eq:l1:sparse:learning} can be formulated as:
\begin{equation}\label{eq:projection}
\min_{\bm \theta: \| X^T  \bm \theta\|_{\infty} \le 1 } \frac{1}{2} \left \| \bm \theta-\frac{\mathbf y}{\lambda} \right \|_2^2.
\end{equation}
For Lasso, the $\lambda_{\max}$ in Eq.~\eqref{eq:lambda:max}
can be analytically computed as $\lambda_{\max} = \|X^T \mathbf y\|_{\infty}$. 
In applying Sasvi, we might start with $\lambda_1=\lambda_{\max}$,
since the primal and dual optimals can be computed analytically as:
$\bm \beta_1^*=\mathbf 0$ and $\bm \theta_1^* = \frac{\mathbf y}{\lambda_{\max}}$.

\subsection{Feasible Set Construction}

Given $\lambda_1$, $\bm \theta_1^*$ and $\lambda_2$, we aim at estimating the upper-bound of $|\langle  \mathbf x_j ,  \bm \theta^*_2 \rangle|$ without the actual computation of $\bm \theta_2^*$. 
To this end, we construct a 
feasible set for $\bm \theta_2^*$, and then estimate the upper-bound  
in the constructed feasible set. 
To construct the feasible set, we make use of the variational inequality that provides
the sufficient and necessary condition of a constrained convex optimization problem.
\begin{lemma}\label{lemma:optimization} \cite{Nesterov:2004}
For the constrained convex optimization problem:
\begin{equation}\label{eq:general:problem}
 \min_ {\mathbf x \in G} f (\mathbf x),
\end{equation}
with $G$ being convex and closed and $f(\cdot)$ being convex and differentiable, $\mathbf x^* \in G$ is an optimal solution
of Eq.~\eqref{eq:general:problem} if and only if
\begin{equation}\label{eq:optimality:condition:vi}
  \langle f'(\mathbf x^*), \mathbf x - \mathbf x^* \rangle  \ge 0, \forall \mathbf x \in G.
\end{equation}
\end{lemma}
Eq.~\eqref{eq:optimality:condition:vi} is the so-called variation inequality for the problem in Eq.~\eqref{eq:general:problem}.
Applying Lemma~\ref{lemma:optimization} to the Lasso dual problem in Eq.~\eqref{eq:projection}, we can represent the optimality conditions for $\bm \theta_1^*$ and $\bm \theta_2^*$ 
using the following two variational inequalities:
\begin{equation}\label{eq:optimality:condition:1}
  \left \langle \bm \theta_1^* - \frac{\mathbf y}{\lambda_1}, \bm \theta - \bm \theta_1^* \right \rangle \ge 0, \forall \bm \theta: \|X^T \bm \theta\|_{\infty} \le 1,
\end{equation}
\begin{equation}\label{eq:optimality:condition:2}
  \left \langle \bm \theta_2^* - \frac{\mathbf y}{\lambda_2}, \bm \theta - \bm \theta_2^* \right \rangle \ge 0, \forall \bm \theta: \|X^T \bm \theta\|_{\infty} \le 1.
\end{equation}
Plugging $\bm \theta= \bm \theta_2^*$ and $\bm \theta= \bm \theta_1^*$ into Eq.~\eqref{eq:optimality:condition:1} and Eq.~\eqref{eq:optimality:condition:2} respectively, we have
\begin{equation} \label{eq:optimality:condition:1:2}
  \left \langle \bm \theta_1^* - \frac{\mathbf y}{\lambda_1}, \bm \theta_2^* - \bm \theta_1^* \right \rangle \ge 0, 
\end{equation}
\begin{equation} \label{eq:optimality:condition:2:1}
  \left \langle \bm \theta_2^* - \frac{\mathbf y}{\lambda_2}, \bm \theta_1^* - \bm \theta_2^* \right \rangle \ge 0.
\end{equation}
With Eq.~\eqref{eq:optimality:condition:1:2} and Eq.~\eqref{eq:optimality:condition:2:1}, we can construct the
following feasible set for $\bm \theta_2^*$ as:
\begin{equation} \label{eq:feasible:set}
    \Omega (\bm \theta_2^*)= 
		\{ \bm \theta: \langle \bm \theta_1^* - \frac{\mathbf y}{\lambda_1}, \bm \theta - \bm \theta_1^* \rangle \ge 0,
                   \langle \bm \theta - \frac{\mathbf y}{\lambda_2}, \bm \theta_1^* - \bm \theta \rangle \ge 0 \}.
\end{equation}

For an illustration of the feasible set, please refer to Figure~\ref{fig:explain_theorem}.
Generally speaking, the closer $\lambda_2$ is to $\lambda_1$, the tighter the feasible set for $\bm \theta_2^*$ is.
In fact, when $\lambda_2$ approaches to $\lambda_1$, $\Omega (\bm \theta_2^*)$ concentrates
to a singleton set that only contains $\bm \theta_2^*$.
Note that one may use additional $\bm \theta$'s
in Eq.~\eqref{eq:optimality:condition:2} for improving the estimation of the feasible set of $\bm \theta_2^*$.
Next, we discuss how to make use of the feasible set defined in Eq.~\eqref{eq:feasible:set}
for estimating an upper-bound for $|\langle  \mathbf x_j ,  \bm \theta^*_2 \rangle|$.

\subsection{Upper-bound Estimation}\label{ss:upper:bound}

Since $\bm \theta_2^* \in \Omega (\bm \theta_2^*)$,
we can estimate an upper-bound of $|\langle  \mathbf x_j ,  \bm \theta^*_2 \rangle| $
by solving
\begin{equation}\label{eq:proposal:II}
\max_{\bm \theta \in \Omega (\bm \theta_2^*)} 
|\langle \mathbf x_j, \bm \theta  \rangle|.
\end{equation}

Next, we show how to solve Eq.~\eqref{eq:proposal:II}. For discussion convenience, we introduce the following
three variables:
\begin{equation} \label{eq:a:b:equation}
\begin{aligned}
\mathbf a  &= \frac{\mathbf y}{\lambda_1} - \bm \theta_1^* = \frac{X \bm \beta_1^*}{\lambda_1},  \\
\mathbf b  &= \frac{\mathbf y}{\lambda_2} - \bm \theta_1^* = \mathbf a +  (\frac{\mathbf y}{\lambda_2} - \frac{\mathbf y}{\lambda_1}),\\
\mathbf r  &= 2 \bm \theta - (\bm \theta_1^* + \frac{\mathbf y}{\lambda_2}),  \\
\end{aligned}
\end{equation}
where $\mathbf a$ denotes the prediction based on $\bm \beta_1^*$ scaled by $\frac{1}{\lambda_1}$,
and $\mathbf b$ is the summation of $\mathbf a$ and the change of the inputs to the dual problem in Eq.~\eqref{eq:projection}
from $\lambda_1$ to $\lambda_2$.

\begin{figure}
\begin{center}
\centerline{\includegraphics[width=0.95\columnwidth]{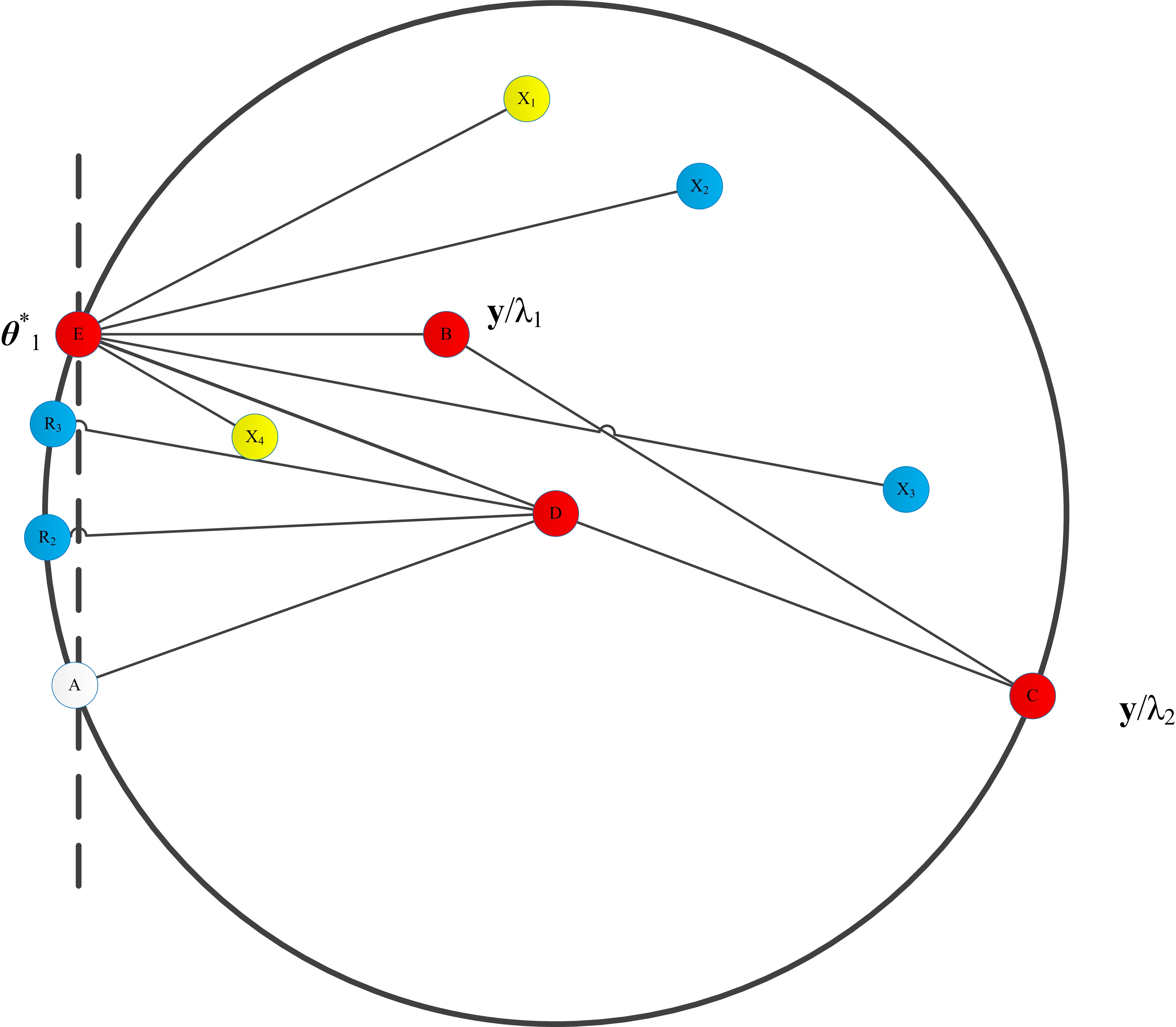}}
\caption{Illustration of the feasible set used in Sasvi and Theorem~\ref{theorem:main:result2}. The points in the figure are explained as follows. 
E:  $\bm \theta_1^*$, 
B: $\frac{\mathbf y}{ \lambda_1}$, 
C: $\frac{\mathbf y}{ \lambda_2}$, 
D: $\frac{\bm \theta_1^* + \frac{\mathbf y}{ \lambda_2}}{2}$. 
The left hand side of the dash line represents the half space $\{\bm \theta: \langle \bm \theta_1^* - \frac{\mathbf y}{\lambda_1}, \bm \theta - \bm \theta_1^* \rangle \ge 0\}$,
and the ball centered at D with radius ED represents $\{\bm \theta: \langle \bm \theta - \frac{\mathbf y}{\lambda_2}, \bm \theta_1^* - \bm \theta \rangle \ge 0 \}$.
For Theorem~\ref{theorem:main:result2}, EX$_1$, EX$_2$, EX$_3$ and EX$_4$ denote $\pm \mathbf x_j$ in two subcases: 1) the angle between EB and EX$_1$ (EX$_4$) is larger than the angle between EB and EC,
and 2) the angle between EB and EX$_2$ (EX$_3$) is smaller than the angle between EB and EC. 
R$_2$ (R$_3$) is the maximizer to Eq.~\eqref{eq:proposal:II} with EX$_2$ (EX$_3$) denoting $\pm \mathbf x_j$.
With EX$_1$  (EX$_4$) denoting $\pm \mathbf x_j$, the maximizer to Eq.~\eqref{eq:proposal:II} 
is on the intersection between the dashed line and the ball centered at D with radius ED.
}\label{fig:explain_theorem}
\end{center}
\vskip -0.1in
\end{figure}

Figure~\ref{fig:explain_theorem} illustrates $\mathbf a$ and $\mathbf b$ by lines EB and EC, respectively. 
For the triangle EBC, the following theorem shows that the angle between $\mathbf a $ and $\mathbf b$ is acute.
\begin{theorem}\label{theorem:main:result1}
Let $\mathbf y \neq \mathbf 0$, and $\|X^T \mathbf y\|_{\infty} \ge \lambda_1 > \lambda_2 >0$. We have
\begin{equation}
     \mathbf b \neq \mathbf 0, \langle \mathbf b, \mathbf a \rangle \ge 0,
\end{equation}
and $\langle \mathbf b, \mathbf a \rangle = 0$ if and only if $\lambda_1 = \|X^T \mathbf y\|_{\infty}$. 
In addition, if $\lambda_1 <\|X^T \mathbf y\|_{\infty}$, then $\mathbf a \neq \mathbf 0$.
\end{theorem}
The proof of Theorem~\ref{theorem:main:result1} is given in Supplement A. 
With the notations in Eq.~\eqref{eq:a:b:equation}, Eq.~\eqref{eq:proposal:II} can be rewritten as
\begin{equation}\label{eq:proposal:II:rewrite}
\begin{aligned}
  \max_{\mathbf r} & \quad \frac{ 1}{2} \left| \left \langle \mathbf x_j,    \bm \theta_1^* + \frac{\mathbf y}{\lambda_2}  \right \rangle +  \langle \mathbf x_j, \mathbf r  \rangle \right| \\
	\mbox{subject to} & \quad   \langle    \mathbf a, \mathbf r + \mathbf b \rangle \le 0,      \|\mathbf r \|_2^2 \le \|\mathbf b \|_2^2.
\end{aligned}
\end{equation}
The objective function of Eq.~\eqref{eq:proposal:II:rewrite} can be represented
by half of the following form:
$$
  \max \left( \langle \mathbf x_j,     \bm \theta_1^* + \frac{\mathbf y}{\lambda_2}  \rangle  +  \langle  \mathbf x_j, \mathbf r  \rangle,  
      - \langle \mathbf x_j,   \bm \theta_1^* + \frac{\mathbf y}{\lambda_2}  \rangle  -  \langle \mathbf x_j, \mathbf r  \rangle \right)
$$
which indicates that Eq.~\eqref{eq:proposal:II:rewrite} can be computed by maximizing $ \langle \mathbf x_j, \mathbf r  \rangle$ and $- \langle \mathbf x_j, \mathbf r  \rangle$ 
over the feasible set in the same equation. Maximizing $\langle \mathbf x_j, \mathbf r  \rangle$ and $- \langle \mathbf x_j, \mathbf r  \rangle$ can
be computed by minimizing $ \langle -\mathbf x_j, \mathbf r  \rangle$ and $  \langle \mathbf x_j, \mathbf r  \rangle$,
which can be solved by the following minimization problem:
\begin{equation} \label{eq:single:problem:study}
\begin{aligned}
  \min_{\mathbf r} & \quad \langle \mathbf x, \mathbf r \rangle \\
	\mbox{subject to} & \quad   \langle   \mathbf a, \mathbf r + \mathbf b \rangle \le 0,   \|\mathbf r \|_2^2 \le \|\mathbf b\|_2^2.
\end{aligned}
\end{equation}
We assume that $\mathbf x$ is a non-zero vector. Let
\begin{equation}\label{eq:x:bot:def}
   \mathbf x^{\bot} = \mathbf x - \mathbf a \langle \mathbf x, \mathbf a \rangle / \|\mathbf a\|_2^2,
\end{equation}
\begin{equation}\label{eq:x:j:bot}
   \mathbf x_j^{\bot} = \mathbf x_j - \mathbf a \langle \mathbf x_j, \mathbf a \rangle /\|\mathbf a\|_2^2,
\end{equation}
\begin{equation}\label{eq:y:bot}
   \mathbf y^{\bot} = \mathbf y - \mathbf a \langle \mathbf y, \mathbf a \rangle / \|\mathbf a\|_2^2,
\end{equation}
which are the orthogonal projections of $\mathbf x$, $\mathbf x_j$, and $\mathbf y$ onto the null space of $\mathbf a$, respectively.
Our next theorem says that Eq.~\eqref{eq:single:problem:study} admits a closed form solution.

\begin{theorem}\label{theorem:min:result}
Let $0< \lambda_1 \le \|X^T \mathbf y\|_{\infty}$, $0<\lambda_2 < \lambda_1$,  $\mathbf x \neq \mathbf 0$ and $\mathbf y \neq \mathbf 0$.
Eq.~\eqref{eq:single:problem:study} equals to $-\|\mathbf x\|_2\|\mathbf b\|_2$, if $\frac{ \langle \mathbf b, \mathbf a \rangle }{ \|\mathbf b \|_2 } \le \frac{\langle \mathbf x , \mathbf a \rangle} { \|\mathbf x\|_2 }$, and $-  \|\mathbf x^{\bot}\|_2 \sqrt{\|\mathbf b\|_2^2 - \frac{ \langle \mathbf b, \mathbf a \rangle ^2}{ \|\mathbf a\|_2^2} } 																			 
	-\frac{ \langle \mathbf a, \mathbf b \rangle \langle \mathbf x, \mathbf a \rangle} { \|\mathbf a\|_2^2}$ otherwise. 
\end{theorem}
 
The proof of Theorem~\ref{theorem:min:result} is given in Supplement B. 
With Theorem~\ref{theorem:min:result}, we can obtain
the upper-bound of $|\langle  \mathbf x_j ,  \bm \theta^*_2 \rangle| $ in the following theorem.
\begin{theorem}\label{theorem:main:result2}
Let $\mathbf y \neq 0$, and $\|X^T \mathbf y\|_{\infty} \ge \lambda_1 > \lambda_2 >0$. 
Denote
\begin{equation}\label{eq:upper:bound:positive}
u_j^+(\lambda_2) =\max_{\bm \theta \in \Omega(\bm \theta_2^*)} 
\langle \mathbf x_j, \bm \theta  \rangle,
\end{equation}
\begin{equation}\label{eq:upper:bound:negative}
u_j^-(\lambda_2) =\max_{\bm \theta \in \Omega(\bm \theta_2^*)} 
\langle - \mathbf  x_j, \bm \theta  \rangle.
\end{equation}

We have:

1) If $ \mathbf a \neq \mathbf 0$ and $\frac{ \langle \mathbf b, \mathbf a \rangle }{\|\mathbf b \|_2  \|\mathbf a \|_2 } > \frac{ | \langle \mathbf x_j , \mathbf a \rangle |} { \|\mathbf x_j\|_2 \|\mathbf a \|_2 }$
then
\begin{equation}\label{eq:bound:little:lambda:positive}
u_j^+(\lambda_2)		=  \langle \mathbf x_j, \bm \theta_1^* \rangle + \frac{\frac{1}{\lambda_2} - \frac{1}{\lambda_1}}{2} \left [ \|\mathbf x_j^{\bot}\|_2 \|\mathbf y^{\bot}\|_2  + \langle \mathbf x_j^{\bot} , \mathbf y ^{\bot} \rangle \right],
\end{equation}
\begin{equation}\label{eq:bound:little:lambda:negative}
u_j^-(\lambda_2)		=  
					    -\langle \mathbf x_j, \bm \theta_1^* \rangle + \frac{\frac{1}{\lambda_2} - \frac{1}{\lambda_1}}{2} \left [ \|\mathbf x_j^{\bot}\|_2 \|\mathbf y^{\bot}\|_2  - \langle \mathbf x_j^{\bot} , \mathbf y ^{\bot} \rangle \right].
\end{equation}

2) If $\langle \mathbf x_j , \mathbf a \rangle >0$ and
$\frac{ \langle \mathbf b, \mathbf a \rangle }{\|\mathbf b \|_2  \|\mathbf a \|_2 } \le \frac{  \langle \mathbf x_j , \mathbf a \rangle  } { \|\mathbf x_j\|_2   \|\mathbf a \|_2 }$, then
$u_j^+(\lambda_2)$ satisfies Eq.~\eqref{eq:bound:little:lambda:positive}, and 
\begin{equation}\label{eq:bound:little:lambda:negative:2}
	 u_j^-(\lambda_2)	 =
	      -\langle \mathbf x_j, \bm \theta_1^* \rangle  +
	      \frac{1}{2} \left[ \|\mathbf x_j\|_2 \|\mathbf b\|_2- \langle \mathbf x_j, \mathbf b \rangle \right] .
\end{equation}
	
3)  If $\langle \mathbf x_j , \mathbf a \rangle <0$ and
$\frac{ \langle \mathbf b, \mathbf a \rangle }{\|\mathbf b \|_2    \|\mathbf a \|_2 } \le \frac{ - \langle \mathbf x_j , \mathbf a \rangle  } { \|\mathbf x_j\|_2   \|\mathbf a \|_2 }$, then
\begin{equation}\label{eq:bound:little:lambda:positive:2}
	   u_j^+(\lambda_2)	=    \langle \mathbf x_j, \bm \theta_1^* \rangle + 
		    \frac{1}{2} \left [\|\mathbf x_j\|_2 \|\mathbf b\|_2 +  \langle \mathbf x_j, \mathbf b \rangle \right]		.
\end{equation}
and $  u_j^-(\lambda_2)	$ satisfies Eq.~\eqref{eq:bound:little:lambda:negative}.

4)  If $ \mathbf a= \mathbf 0$, then
Eq.~\eqref{eq:bound:little:lambda:negative:2} and Eq.~\eqref{eq:bound:little:lambda:positive:2} hold. 

\end{theorem}

The proof of Theorem~\ref{theorem:main:result2} is given in Supplement C. 
An illustration of Theorem~\ref{theorem:main:result2} for different cases can be found 
in Figure~\ref{fig:explain_theorem}. It follows from Eq.~\eqref{eq:simple:result} that, if 
$u_j^+(\lambda_2) <1$ and $u_j^-(\lambda_2) <1$,
then the $j$-th feature can be safely eliminated for the computation of $\bm \beta_2^*$.
We provide the following analysis to the established upper-bound. Firstly,  we have
$$\lim_{\lambda_2 \rightarrow \lambda_1} u_j^+(\lambda_2) =
\langle \mathbf x_j, \bm \theta_1^* \rangle, \lim_{\lambda_2 \rightarrow \lambda_1} u_j^-(\lambda_2) =
-\langle \mathbf x_j, \bm \theta_1^* \rangle, $$
which attributes to the fact that $\lim_{\lambda_2 \rightarrow \lambda_1} \Omega(\bm \theta_2^*) =\{\bm \theta_1^*\}$.
Secondly, in the extreme case that $\mathbf x_j$ is orthogonal to 
the scaled prediction $\mathbf a= \frac{ X \bm \beta_1^*}{\lambda_1}$ which is nonzero, Theorem~\ref{theorem:main:result2} 
leads to $\mathbf x_j^{\bot}=\mathbf 0$, $u_j^+(\lambda_2)		=  \langle \mathbf x_j, \bm \theta_1^* \rangle$ 
and $u_j^-(\lambda_2)		=  -\langle \mathbf x_j, \bm \theta_1^* \rangle$. Thus, the $j$-th feature can be safely 
removed for any positive $\lambda_2$ that is smaller than $\lambda_1$ so long as $ |\langle \mathbf x_j, \bm \theta_1^* \rangle| <1$.
Thirdly, in the case that $\mathbf x_j$ has low correlation with
the prediction $\mathbf a= \frac{ X \bm \beta_1^*}{\lambda_1}$, Theorem~\ref{theorem:main:result2}
indicates that the $j$-th feature is very likely to be safely removed for a wide range of $\lambda_2$
if $|\langle \mathbf x_j, \bm \theta_1^* \rangle| <1$.
The monotone properties of the upper-bound established in Theorem~\ref{theorem:main:result2} is given Section~\ref{s:sure:remove}.

%
%

%% file: relationship.tex
Our proposed Sasvi differs from the existing screening techniques~\cite{Ghaoui:2012,Tibshirani:2012,Wang:2012:report,Xiang:2011}
in the construction of the feasible set for $\bm \theta_2^*$.

\subsection{Comparison with the Strong Rule}
The strong rule~\cite{Tibshirani:2012} works on $0< \lambda_2 < \lambda_1$ and makes use of the assumption
\begin{equation}\label{eq:strong:assum}
  | \lambda_2  \langle \mathbf x_j, \bm \theta_2^* \rangle- \lambda_1  \langle \mathbf x_j, \bm \theta_1^* \rangle | \le |\lambda_2 - \lambda_1|,
\end{equation}
from which we can obtain an estimated upper-bound for $|\langle \mathbf x_j, \bm \theta_2^* \rangle| $ as:
\begin{equation}\label{eq:strong:bound}
\begin{aligned}
  |\langle \mathbf x_j, \bm \theta_2^* \rangle| 
	    & \le \frac{|\lambda_1 \langle \mathbf x_j, \bm \theta_1^* \rangle | + | \lambda_2 \langle \mathbf x_j, \bm \theta_2^* \rangle - \lambda_1  \langle \mathbf x_j, \bm \theta_1^* \rangle |}{\lambda_2 } \\
	    & \le \frac{|\lambda_1 \langle \mathbf x_j, \bm \theta_1^* \rangle | + (\lambda_1 - \lambda_2)}{\lambda_2}\\
			& = \frac{\lambda_1}{\lambda_2} | \langle \mathbf x_j, \bm \theta_1^* \rangle | + \left[ \frac{\lambda_1}{\lambda_2} - 1 \right]
\end{aligned}
\end{equation}
A comparison between Eq.~\eqref{eq:strong:bound} and the upper-bound established in Theorem~\ref{theorem:main:result2} shows
that, 1) both are dependent on $\langle \mathbf x_j, \bm \theta_1^* \rangle $, the inner product between the $j$-th feature and
the dual variable $\bm \theta_1^*$ obtained at $\lambda_1$, but note that $\frac{\lambda_1}{\lambda_2}>1$, 2) in comparison with the data independent term $\frac{\lambda_1}{\lambda_2} - 1 $ used in the strong rule, Sasvi utilizes
a data dependent term as shown in Eqs.~\eqref{eq:bound:little:lambda:positive}-\eqref{eq:bound:little:lambda:positive:2}.
We note that, 1) when a feature $\mathbf x_j$ has low correlation with
the prediction $\mathbf a= \frac{ X \bm \beta_1^*}{\lambda_1}$, the upper-bound for $| \langle \mathbf x_j, \bm \theta_2^* \rangle|$ 
estimated by Sasvi might
be lower than the one by the strong rule~\footnote{
According to the analysis given at the end of Section~\ref{ss:upper:bound},
this argument is true for the extreme case that $\mathbf x_j$ is orthogonal to 
the nonzero prediction $\mathbf a= \frac{ X \bm \beta_1^*}{\lambda_1}$.}, and 2) as pointed out in~\cite{Tibshirani:2012}, Eq.~\eqref{eq:strong:assum} might not always hold, and 
the same applies to Eq.~\eqref{eq:strong:bound}.

\begin{figure}
\vskip 0.2in
\begin{center}
\centerline{\includegraphics[width=0.95\columnwidth]{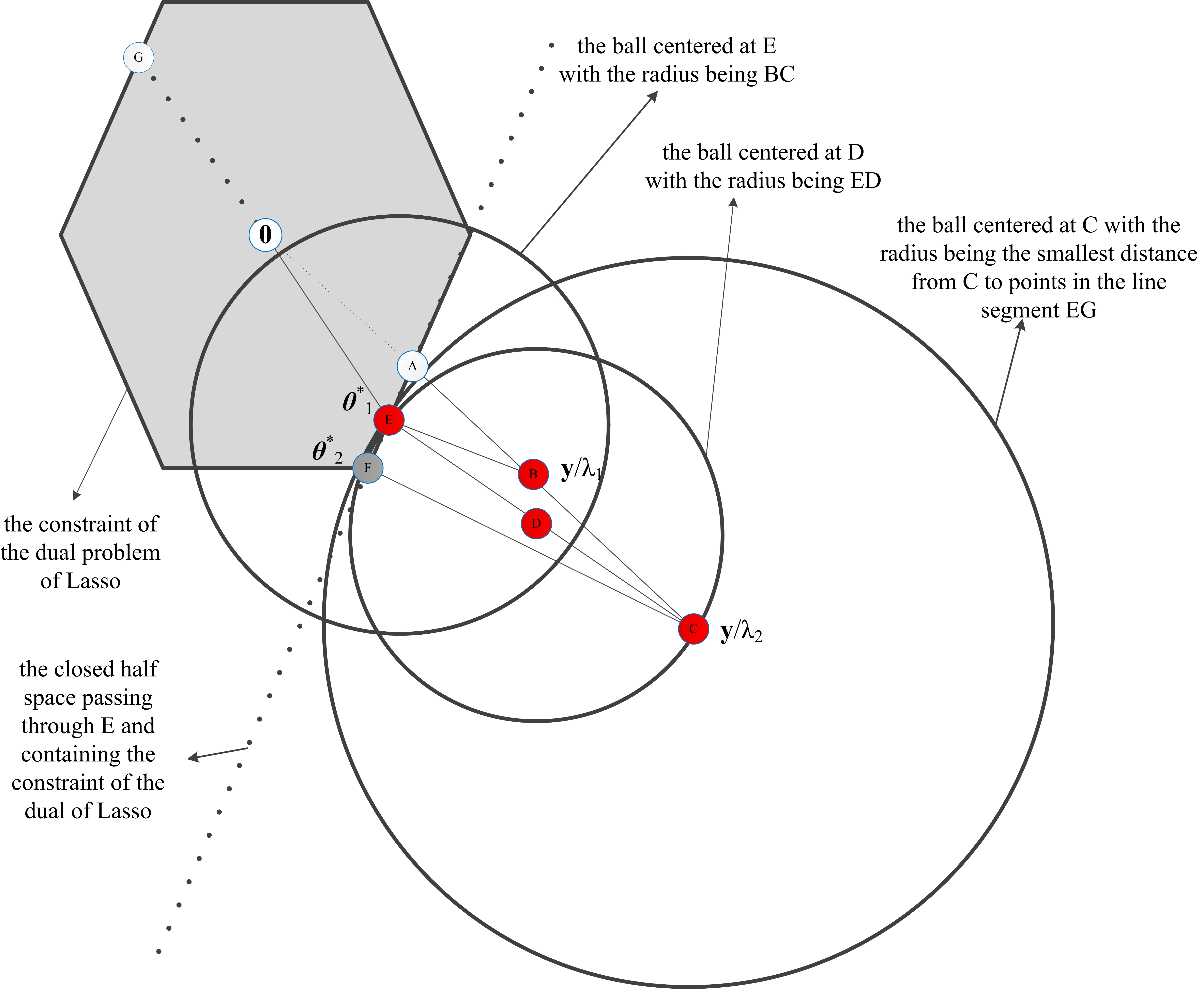}}
\caption{Comparison of Sasvi with existing safe screening approaches. The points in the figure are as follows. A: $\frac{\mathbf y}{ \lambda_{\max}}$, B: $\frac{\mathbf y}{ \lambda_1}$, C: $\frac{\mathbf y}{ \lambda_2}$, D: the middle point of C and E,
E: $\bm \theta_1^*$, F: $\bm \theta_2^*$, and G: $-\bm \theta_1^*$.
The feasible set for $\bm \theta_2^*$ used by the proposed Sasvi approach is the intersection between the ball centered at D with radius being half EC and the closed half space passing through E and containing the constraint of the dual of Lasso.
The feasible set for $\bm \theta_2^*$ used by the SAFE~\cite{Ghaoui:2012} approach is the ball centered at C with radius being the smallest distance from C to the points in the line segment EG.
The feasible set for $\bm \theta_2^*$ used by the DPP~\cite{Wang:2012:report} approach is the ball centered at E with radius BC.
 }\label{fig:comparison}
\end{center}
\vskip -0.2in
\end{figure}

Next, we compare Sasvi with the SAFE approach~\cite{Ghaoui:2012} and the DPP approach~\cite{Wang:2012:report}, and the 
differences in terms of the feasible sets are shown in Figure~\ref{fig:comparison}.

\subsection{Comparison with the SAFE approach}

Denote $ G(\bm \theta)  =  \frac{1}{2} \|  \mathbf y ||_2^2 - \frac{1}{2} \| \lambda_2 \bm \theta- \mathbf y ||_2^2$.
The SAFE approach makes use of the so-called ``dual" scaling, and compute the upper-bound of the $ G(\bm \theta)$ for  $\lambda_2 $ as 
\begin{equation}\label{eq:safe:subproblem}
   \gamma(\lambda_2) = \max_{s: |s|\le 1 } G( s \bm \theta) = \max_{s: |s|\le 1 } \frac{1}{2} \|  \mathbf y ||_2^2 - \frac{1}{2} \| s \lambda_2 \bm \theta_1^*- \mathbf y ||_2^2,
\end{equation}
Note that, compared to the SAFE paper, the dual variable $\bm \theta$ has been scaled in the formulation in Eq.~\eqref{eq:safe:subproblem}, 
but this scaling does not influence of the following result for the SAFE approach. Denote $s^*$ as the optimal solution.
Solving Eq.~\eqref{eq:safe:subproblem}, we have $s^* =\max \left( \min \left(  \frac{\langle \bm \theta_1^*, \mathbf y \rangle }{\lambda_2 \|\bm \theta_1^*\|_2}, 1 \right), -1 \right)$ when $\bm \theta_1 \neq \mathbf 0$. 
The SAFE approach 
computes the upper-bound for $|\langle \mathbf x_j, \bm \theta_2^* \rangle|$ as follows:
\begin{equation}\label{eq:safe:bound}
\begin{aligned}
   |\langle \mathbf x_j, \bm \theta_2^* \rangle| 
	 & \le  \max_{\bm \theta: G(\bm \theta) \ge  \gamma(\lambda_2)}  |\langle \mathbf x_j, \bm \theta  \rangle| \\
	 & =  \max_{\bm \theta:   \|  \bm \theta- \frac{\mathbf y}{\lambda_2} ||_2  \le   \| s^* \bm \theta_1^*- \frac{\mathbf y}{\lambda_2} ||_2 }  |\langle \mathbf x_j, \bm \theta  \rangle| \\
 	 & = \frac{|\langle \mathbf x_j, \mathbf y \rangle|}{\lambda_2} + \|\mathbf x_j \|_2  \left\| s^* \bm \theta_1^*- \frac{\mathbf y}{\lambda_2} \right\|_2.
\end{aligned}
\end{equation}
Next, we show that the feasible set for $\bm \theta_2^*$ used in Eq.~\eqref{eq:safe:bound} can
be derived from the variational inequality in Eq.~\eqref{eq:optimality:condition:2} followed by
relaxations.

Utilizing  $\|X^T \bm \theta_1^*\|_{\infty} \le 1$ and $ |s^*| \le 1$,
we can set $\bm \theta = s^* \bm \theta_1^*$ in Eq.~\eqref{eq:optimality:condition:2} and obtain
\begin{equation}\label{eq:optimality:condition:2:safe}
  \left \langle \bm \theta_2^* - \frac{\mathbf y}{\lambda_2}, s^* \bm \theta_1^* - \bm \theta_2^*  \right \rangle \ge 0,
\end{equation}
which leads to
\begin{equation}\label{eq:optimality:condition:2:safe:1}
\begin{aligned}
& \left \langle \bm \theta_2^* - \frac{\mathbf y}{\lambda_2},  \bm \theta_2^* - \frac{\mathbf y}{\lambda_2}  \right \rangle  
    - \left \langle \bm \theta_2^* - \frac{\mathbf y}{\lambda_2},  s^* \bm \theta_1^*  -\frac{\mathbf y}{\lambda_2}  \right  \rangle \\
& =  \left \langle \bm \theta_2^* - \frac{\mathbf y}{\lambda_2},  \bm \theta_2^* - \frac{\mathbf y}{\lambda_2}  +  \frac{\mathbf y}{\lambda_2}  - s^* \bm \theta_1^*   \right \rangle
	\le 0.
\end{aligned}
\end{equation}
Since 
\begin{equation}\label{eq:relax:111}
   \left \langle \bm \theta_2^* - \frac{\mathbf y}{\lambda_2},  s^* \bm \theta_1^*  -\frac{\mathbf y}{\lambda_2}   \right  \rangle \le
\left \|\bm \theta_2^* - \frac{\mathbf y}{\lambda_2} \right \|_2 
\left \| s^* \bm \theta_1^*  -\frac{\mathbf y}{\lambda_2} \right \|_2, 
\end{equation}
we have
\begin{equation}\label{eq:optimality:condition:2:safe:2}
  \left\|\bm \theta_2^* - \frac{\mathbf y}{\lambda_2} \right \|_2  \le \left \| s^* \bm \theta_1^*  -\frac{\mathbf y}{\lambda_2}   \right\|_2,
\end{equation}
which is the feasible set used in Eq.~\eqref{eq:safe:bound}. 
Note that, the ball defined by Eq.~\eqref{eq:optimality:condition:2:safe:2} 
has higher volume than the one defined by Eq.~\eqref{eq:optimality:condition:2:safe}
due to the relaxation used in Eq.~\eqref{eq:relax:111}, and it can be shown that
the ball defined by Eq.~\eqref{eq:optimality:condition:2:safe} lies within
the ball defined by Eq.~\eqref{eq:optimality:condition:2:safe:2}.


\subsection{Comparison with the DPP approach}

The feasible set for $\bm \theta_2^*$ used in the DPP approach is
\begin{equation} \label{eq:example:12}
  \left \| \frac{\mathbf y}{\lambda_2} - \frac{\mathbf y}{\lambda_1} \right\|_2 \ge  \left \|\bm \theta_2^* - \bm \theta_1^*  \right \|_2, 
\end{equation}
which can be obtained by
\begin{equation} \label{eq:example:11}
  \left \langle\frac{\mathbf y}{\lambda_2} - \frac{\mathbf y}{\lambda_1}, \bm \theta_2^* - \bm \theta_1^* \right \rangle \ge \langle \bm \theta_2^* - \bm \theta_1^* , \bm \theta_2^* - \bm \theta_1^* \rangle.
\end{equation}
and
\begin{equation}\label{eq:example:33}
   \left \langle\frac{\mathbf y}{\lambda_2} - \frac{\mathbf y}{\lambda_1}, \bm \theta_2^* - \bm \theta_1^*  \right \rangle \le \left \|  \frac{\mathbf y}{\lambda_2} - \frac{\mathbf y}{\lambda_1} \right \|_2 \| \bm \theta_2^* - \bm \theta_1^*   \|_2,
\end{equation}
where Eq.~\eqref{eq:example:11} is a result of adding Eq.~\eqref{eq:optimality:condition:1:2} and Eq.~\eqref{eq:optimality:condition:2:1}.
Therefore, 
although the authors in~\cite{Wang:2012:report} motivates the DPP approach from the viewpoint of Euclidean projection, the DPP approach can indeed be treated as generating the 
feasible set for $\bm \theta_2^*$ using the variational inequality in Eq.~\eqref{eq:optimality:condition:1} and Eq.~\eqref{eq:optimality:condition:2} followed by relaxation in Eq.~\eqref{eq:example:33}. 
Note that, the ball specified by Eq.~\eqref{eq:example:12} has higher volume than the one specified by Eq.~\eqref{eq:example:11} due to the relaxation used in Eq.~\eqref{eq:example:33}, and it can be shown that
the ball defined by Eq.~\eqref{eq:example:11} lies within
the ball defined by Eq.~\eqref{eq:example:12}.

%% file: sureRemove.tex
In this subsection, we study the monotone properties of the upper-bound 
established in Theorem~\ref{theorem:main:result2} with regard to the regularization parameter $\lambda_2$.
With such study, we can identify the feature sure removal parameter---the smallest value of $\lambda$ above which a feature is guaranteed  
to have zero coefficient and thus can be safely removed.

Without loss of generality, we assume $ \langle \mathbf x_j, \mathbf a \rangle \ge 0$ and the results can
be easily extended to the case  $ \langle \mathbf x_j, \mathbf a \rangle < 0$.
In addition, we assume that  if $\lambda_1 \neq \|X^T \mathbf y\|_{\infty}$ then
$\bm \theta_1^* \neq \frac{\mathbf y}{\|X^T \mathbf y\|_{\infty}}$. This is a valid assumption for real data.

\begin{figure}[h]
\vskip 0.1in
\begin{center}
\includegraphics[width=0.42\columnwidth]{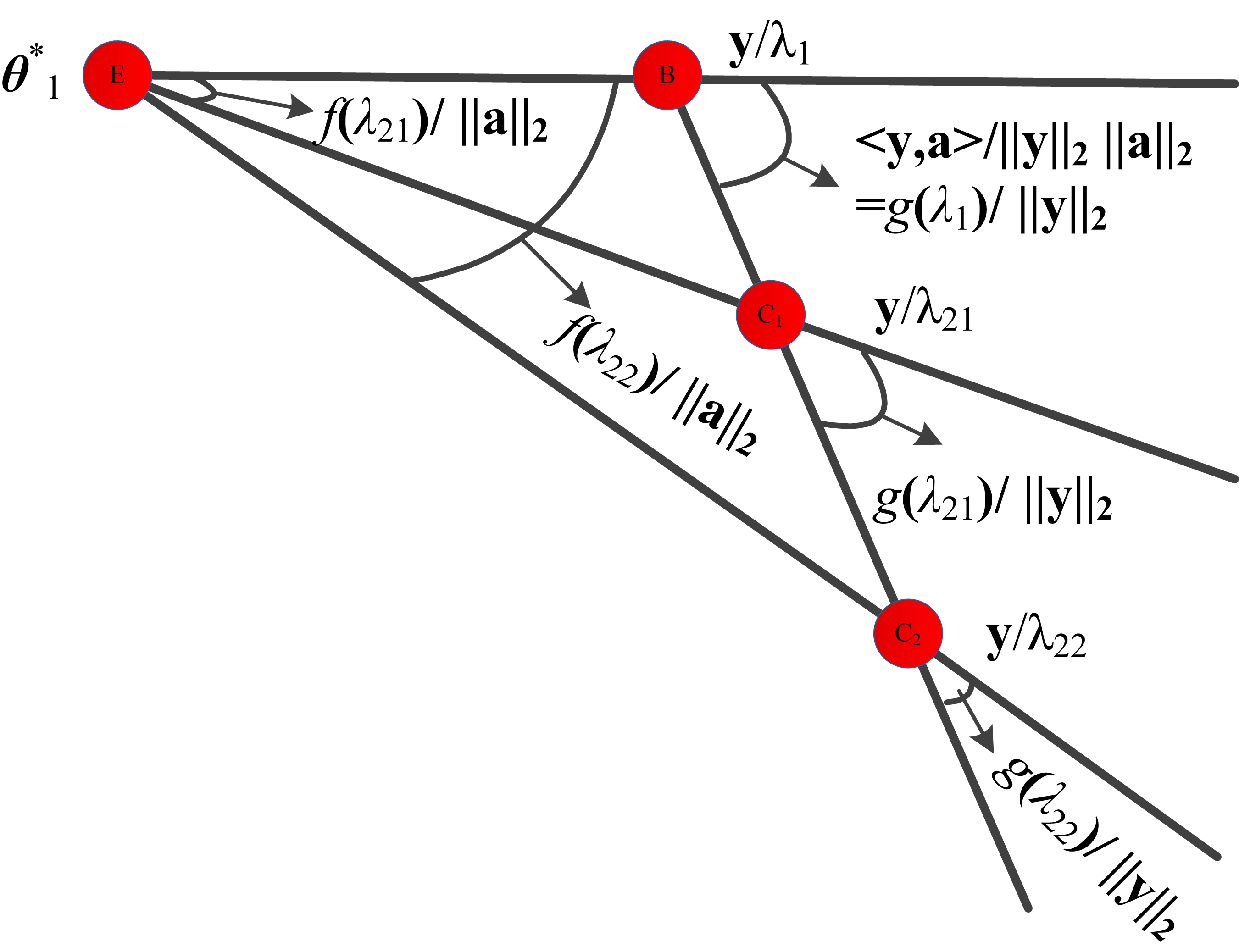}
\includegraphics[width=0.42\columnwidth]{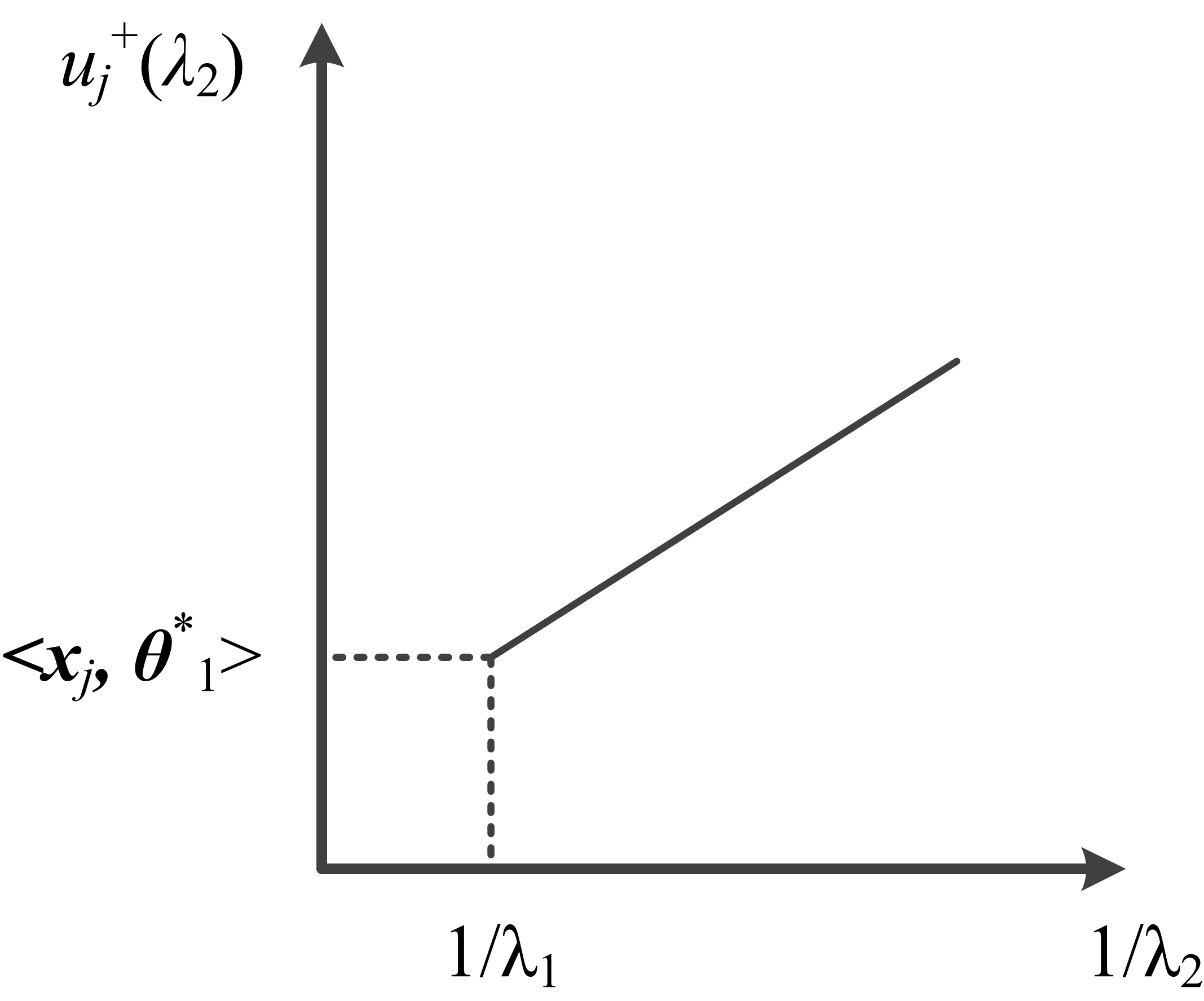}
\includegraphics[width=0.42\columnwidth]{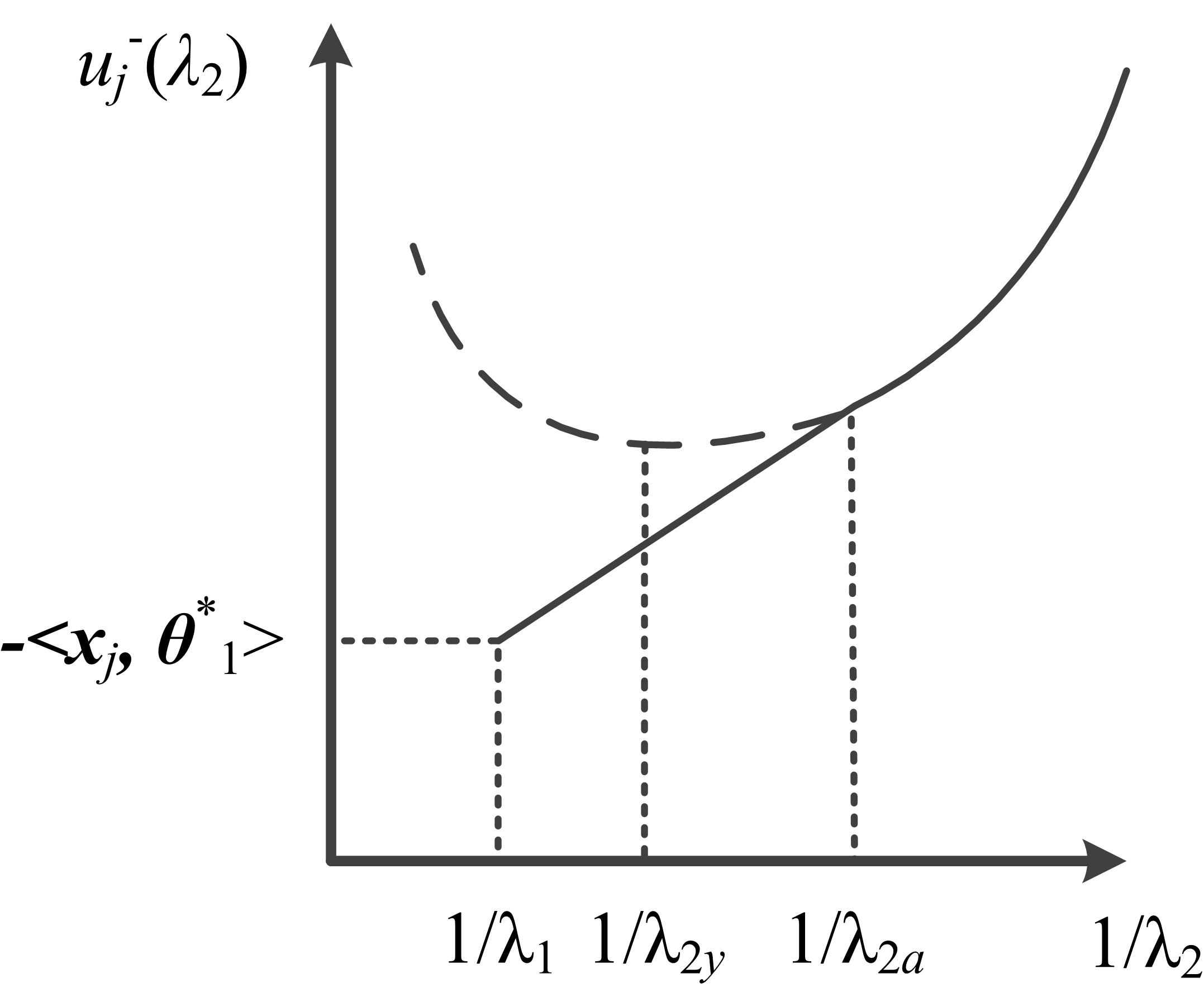}
\includegraphics[width=0.42\columnwidth]{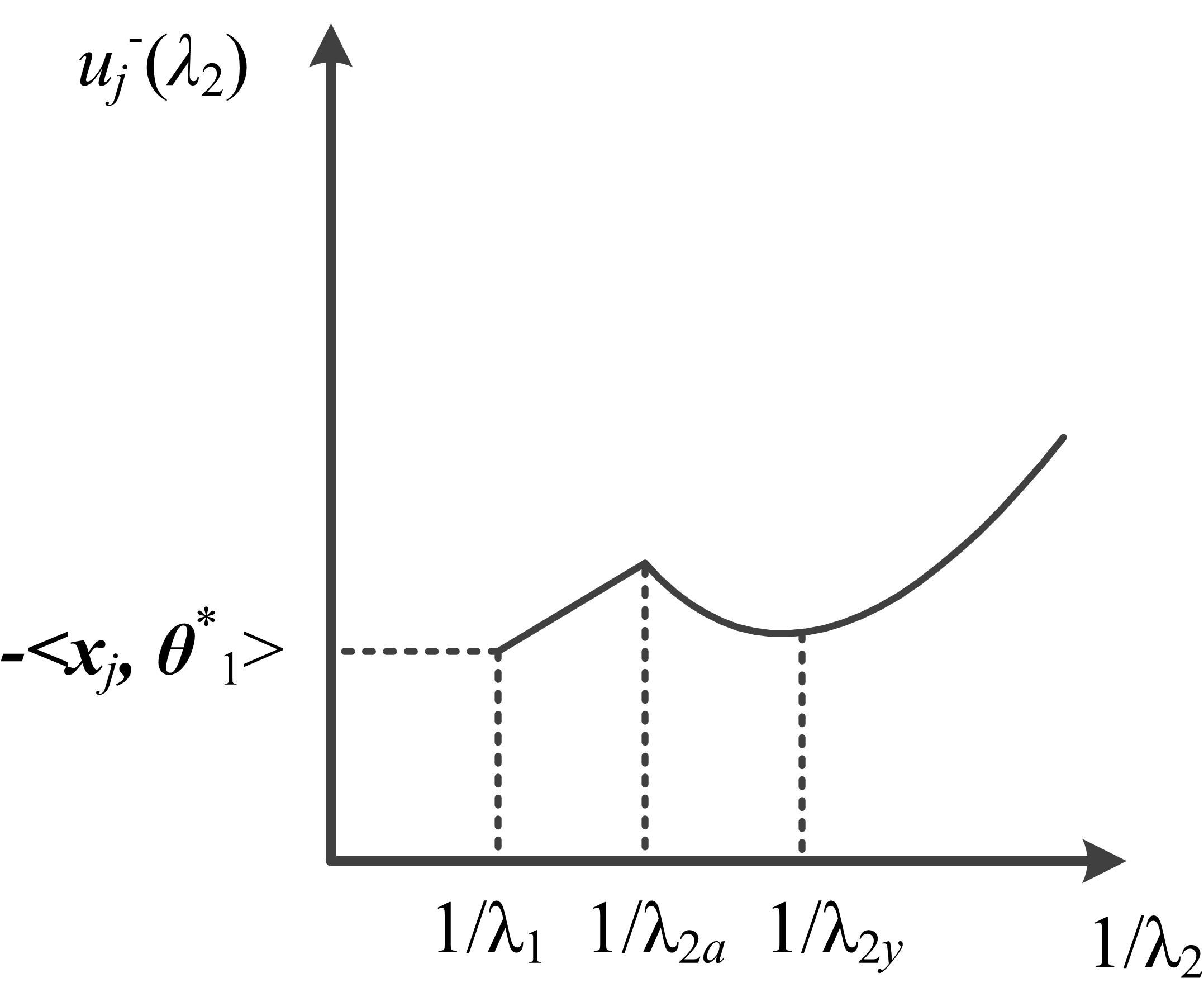}
\caption{Illustration of the monotone properties of Sasvi for Lasso with the assumption $ \langle \mathbf x_j, \mathbf a \rangle \ge 0$.
The first plot geometrically shows the monotone properties of $f(\lambda)$ and $g(\lambda)$, respectively.
The last three plots correspond to the three cases in Theorem~\ref{theorem:main:result3}.
For illustration convenience, the x-axis denotes $\frac{1}{\lambda_2}$ rather than $\lambda_2$.
}\label{fig:sureRemove}
\end{center}
\vskip -0.1in
\end{figure}

Let $\mathbf y \neq 0$, and $\lambda_{\max}=\|X^T \mathbf y\|_{\infty} \ge \lambda_1 \ge \lambda >0$\footnote{If $\lambda_1 \ge \lambda_{\max}$, we have
$\bm \beta_1^*=\mathbf 0$ and thus we focus on $\lambda_1 \le \lambda_{\max}$. In addition, for given $\lambda_1$, we are interested in the screening
for a smaller regularization parameter, i.e., $\lambda < \lambda_1$.}. 
We introduce the following two auxiliary functions:
\begin{equation}\label{eq:aux:f}
  f(\lambda) =  \frac{ \langle  \frac{\mathbf y}{\lambda} - \bm \theta_1^*    , \mathbf a \rangle }{\|  \frac{\mathbf y}{\lambda} - \bm \theta_1^* \|_2  }
\end{equation}
\begin{equation}\label{eq:aux:g}
  g(\lambda) =  \frac{ \langle  \frac{\mathbf y}{\lambda} - \bm \theta_1^*    , \mathbf y \rangle }{\|  \frac{\mathbf y}{\lambda} - \bm \theta_1^* \|_2  }
\end{equation}
We show in Supplement D 
that $f(\lambda)$ is 
strictly increasing with regard to $\lambda$ in $(0,  \lambda_1 ]$ and 
$g(\lambda)$ is strictly decreasing with regard to $\lambda$ in $(0,  \lambda_1 ]$. 
Such monotone properties, which are illustrated geometrically in the first plot of Figure~\ref{fig:sureRemove},
guarantee that $f(\lambda)= \frac{ \langle \mathbf x_j, \mathbf a \rangle }{\|\mathbf x_j\|_2  }$
and $g(\lambda)=  \frac{ \langle \mathbf x_j, \mathbf y \rangle }{\|\mathbf x_j\|_2  }$
have unique roots with regard to $\lambda$ when some conditions are satisfied.

Our main results are summarized in the following theorem:
\begin{theorem}\label{theorem:main:result3}
Let $\mathbf y \neq 0$ and $\|X^T \mathbf y\|_{\infty} \ge \lambda_1 > \lambda_2 >0$. 
Let $ \langle \mathbf x_j, \mathbf a \rangle \ge 0$.
Assume that if $\lambda_1 \neq \|X^T \mathbf y\|_{\infty}$ then $\bm \theta_1^* \neq \frac{\mathbf y}{\|X^T \mathbf y\|_{\infty}}$.

Define $\lambda_{2,a}$ as follows: If $ \frac{ \langle \mathbf y, \mathbf a \rangle }{\|\mathbf y\|_2  }  \ge \frac{ \langle \mathbf x_j, \mathbf a \rangle }{\|\mathbf x_j\|_2  }$, 
then let $\lambda_{2,a}=0$; otherwise, let $\lambda_{2,a}$ be the unique value in $(0, \lambda_1]$
that satisfies  $f(\lambda_{2,a})= \frac{ \langle \mathbf x_j, \mathbf a \rangle }{\|\mathbf x_j\|_2  }$.

Define $\lambda_{2,y}$ as follows: If $\mathbf a=\mathbf 0$ or if $\mathbf a \neq \mathbf 0$ and $ \frac{ \langle \mathbf a, \mathbf y \rangle }{\|\mathbf a\|_2   }  \ge \frac{ \langle \mathbf x_j, \mathbf y \rangle }{\|\mathbf x_j\|_2  }$, 
then let $\lambda_{2,y}=\lambda_1$; otherwise, let $\lambda_{2,y}$ be the unique value in $(0, \lambda_1]$
that satisfies $g(\lambda_{2,y})= \frac{ \langle \mathbf x_j, \mathbf y \rangle }{\|\mathbf x_j\|_2 }$.

We have the following monotone properties:
\begin{itemize}
  \item [1.] $u_j^+ (\lambda_2)$ is monotonically decreasing with regard to  $\lambda_2$ in $(0, \lambda_1]$.
	
	\item [2.] If $\lambda_{2,a} \le \lambda_{2,y}$, then $u_j^- (\lambda_2)$ is monotonically decreasing with regard to
	$\lambda_2$ in $(0, \lambda_1]$.
	
	\item [3.] If $\lambda_{2,a} > \lambda_{2,y}$, then $u_j^- (\lambda_2)$  is monotonically decreasing with regard to 
	$\lambda_2$ in $(0, \lambda_{2,y}) $ and $(\lambda_{2,a}, \lambda_1) $, but monotonically increasing with regard to $\lambda_2$ in $[\lambda_{2,y}, \lambda_{2,a}]$.
\end{itemize}

\end{theorem}

The proof of Theorem~\ref{theorem:main:result3} is given in Supplement D. 
Note that, $\lambda_{2,a}$ and $\lambda_{2,y}$  are dependent on the index $j$,
which is omitted for discussion convenience. 
Figure~\ref{fig:sureRemove} illustrates results presented in Theorem~\ref{theorem:main:result3}. 
The first two cases of Theorem~\ref{theorem:main:result3} indicate that,
if the $j$-th feature $\mathbf x_j$ can be safely removed for a regularization parameter $\lambda=\lambda_2$, then
it can also be safely discarded for any regularization parameter $\lambda $ larger than $\lambda_2$. However, 
the third case in Theorem~\ref{theorem:main:result3} says that this is not always true.
This somehow coincides with the characteristic of Lasso that, a feature that is inactive 
for a regularization parameter $\lambda=\lambda_2$ might become active for a larger
regularization parameter $\lambda > \lambda_2$. In other words, when following the Lasso solution path with a decreasing
regularization parameter, a feature that enters into the model might get removed.

By using Theorem~\ref{theorem:main:result3}, we can 
easily identify for each feature a sure removable parameter $\lambda_s$ that satisfies 
$u_j^+ (\lambda) <1$ and $u_j^- (\lambda) <1$, $\forall \lambda > \lambda_s$.
Note that Theorem~\ref{theorem:main:result3} assumes $ \langle \mathbf x_j, \mathbf a \rangle \ge 0$,
but it can be easily extended to 
the case $ \langle \mathbf x_j, \mathbf a \rangle < 0$ by replacing $\mathbf x_j$ with $-\mathbf x_j$.

%% file: experiment2.tex
In this section, we conduct experiments to evaluate the performance of the proposed Sasvi
in comparison with the sequential SAFE rule \cite{Ghaoui:2012}, the sequential strong rule~\cite{Tibshirani:2012}, and the sequential DPP \cite{Wang:2012:report}.
Note that, SAFE, Sasvi and DPP methods are ``safe" in the sense that the discarded features are guaranteed to have 0 coefficients in the true solution, 
and the strong rule---which is a heuristic rule---might make error and such error was corrected by a KKT condition check as suggested in~\cite{Tibshirani:2012}.

\begin{figure}
\centering
\includegraphics[width=0.42\columnwidth]{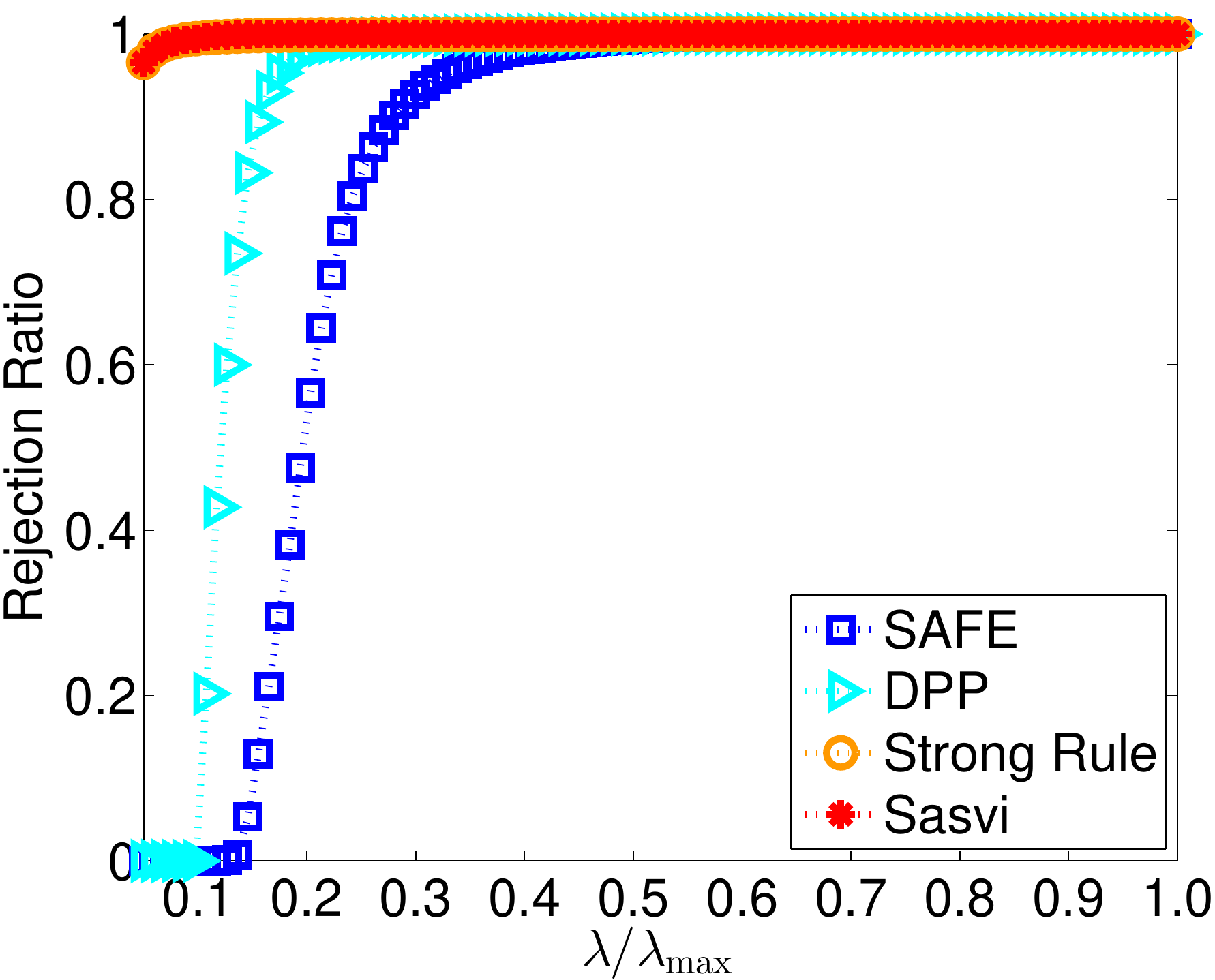}
\includegraphics[width=0.42\columnwidth]{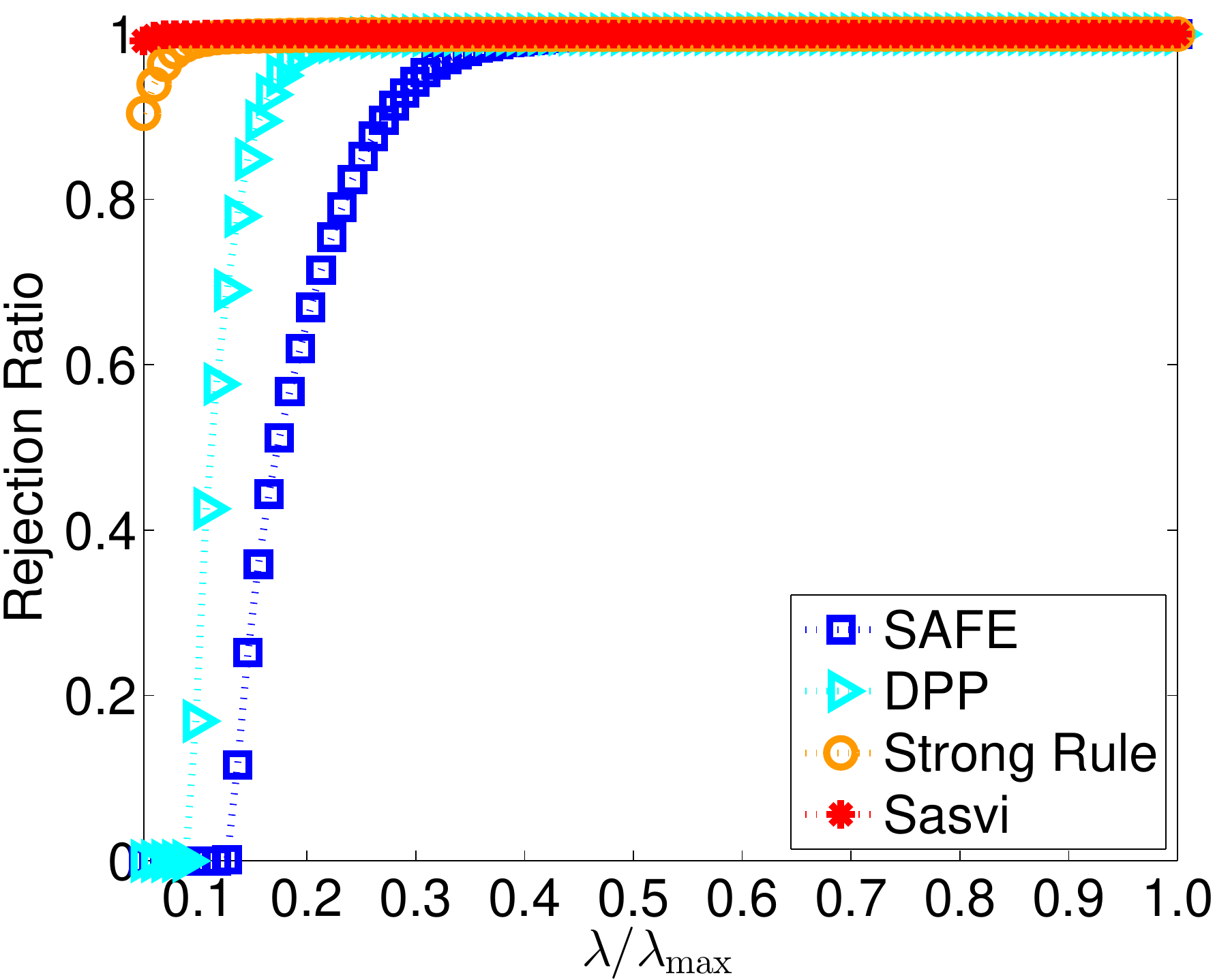} \\
(Real: MNIST) \hspace{0.42in} (Real: PIE)\\
\includegraphics[width=0.42\columnwidth]{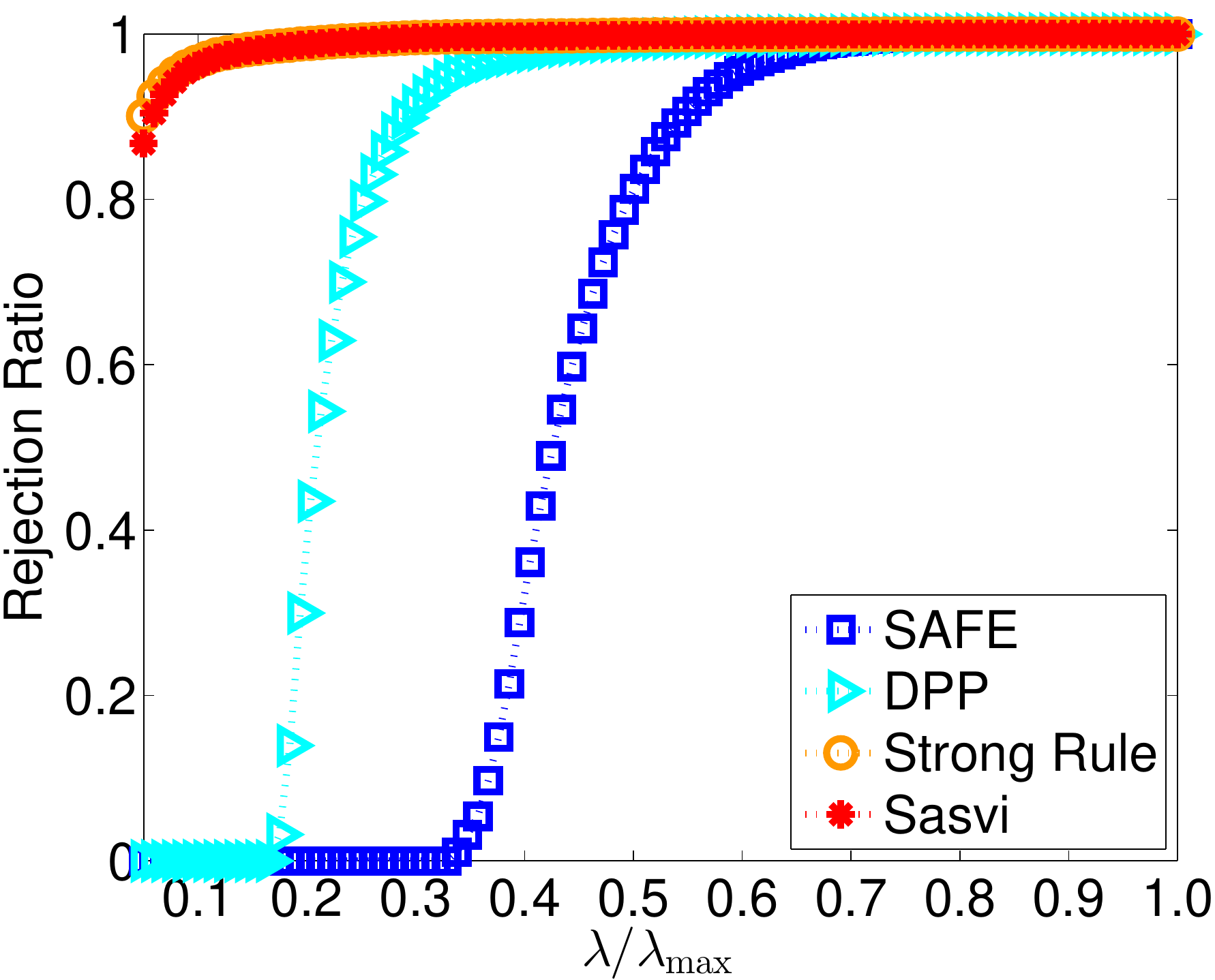}
\includegraphics[width=0.42\columnwidth]{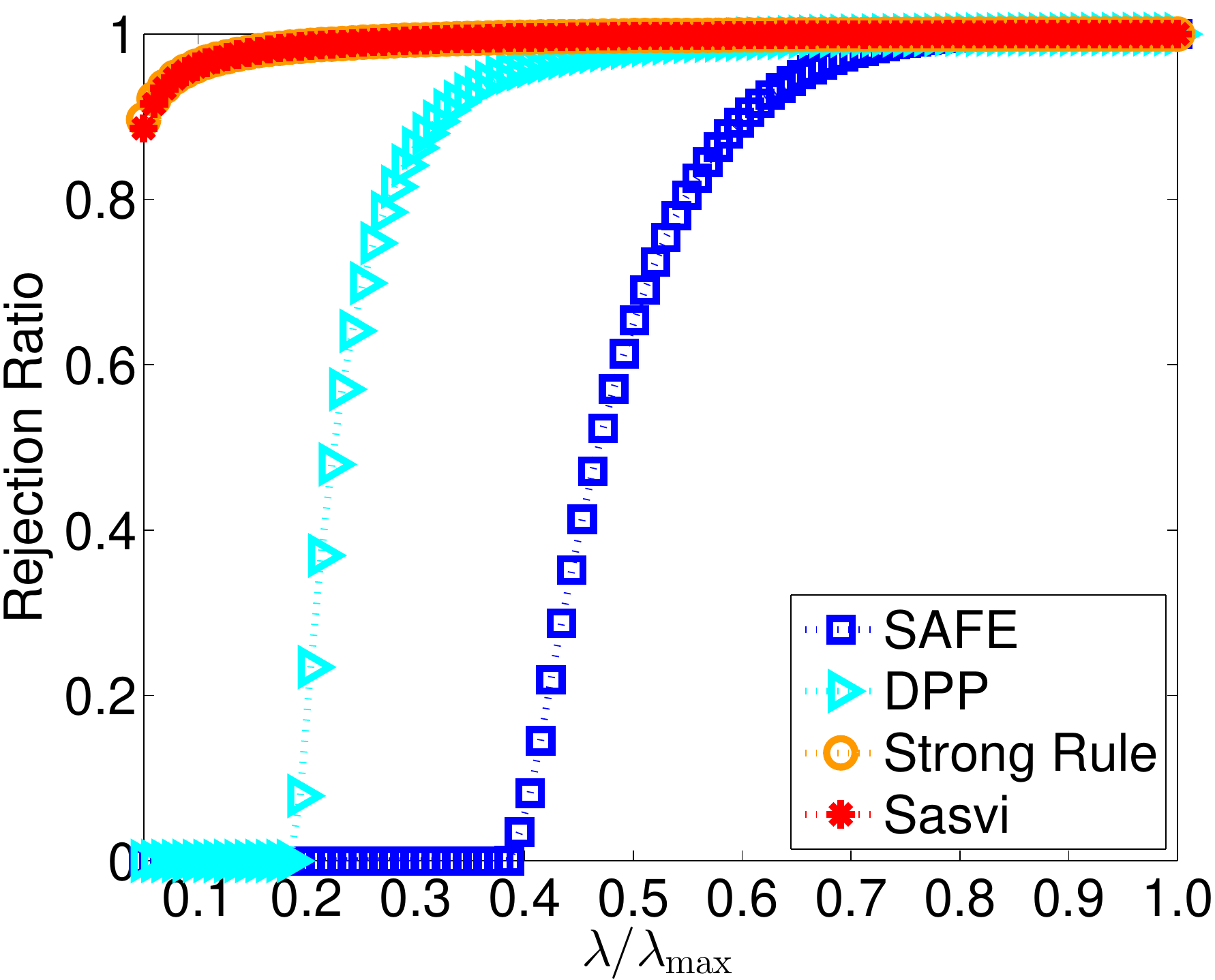}\\
(Synthetic, $\bar p=100$) \hspace{0.15in} (Synthetic, $\bar p=1000$) \\
\includegraphics[width=0.42\columnwidth]{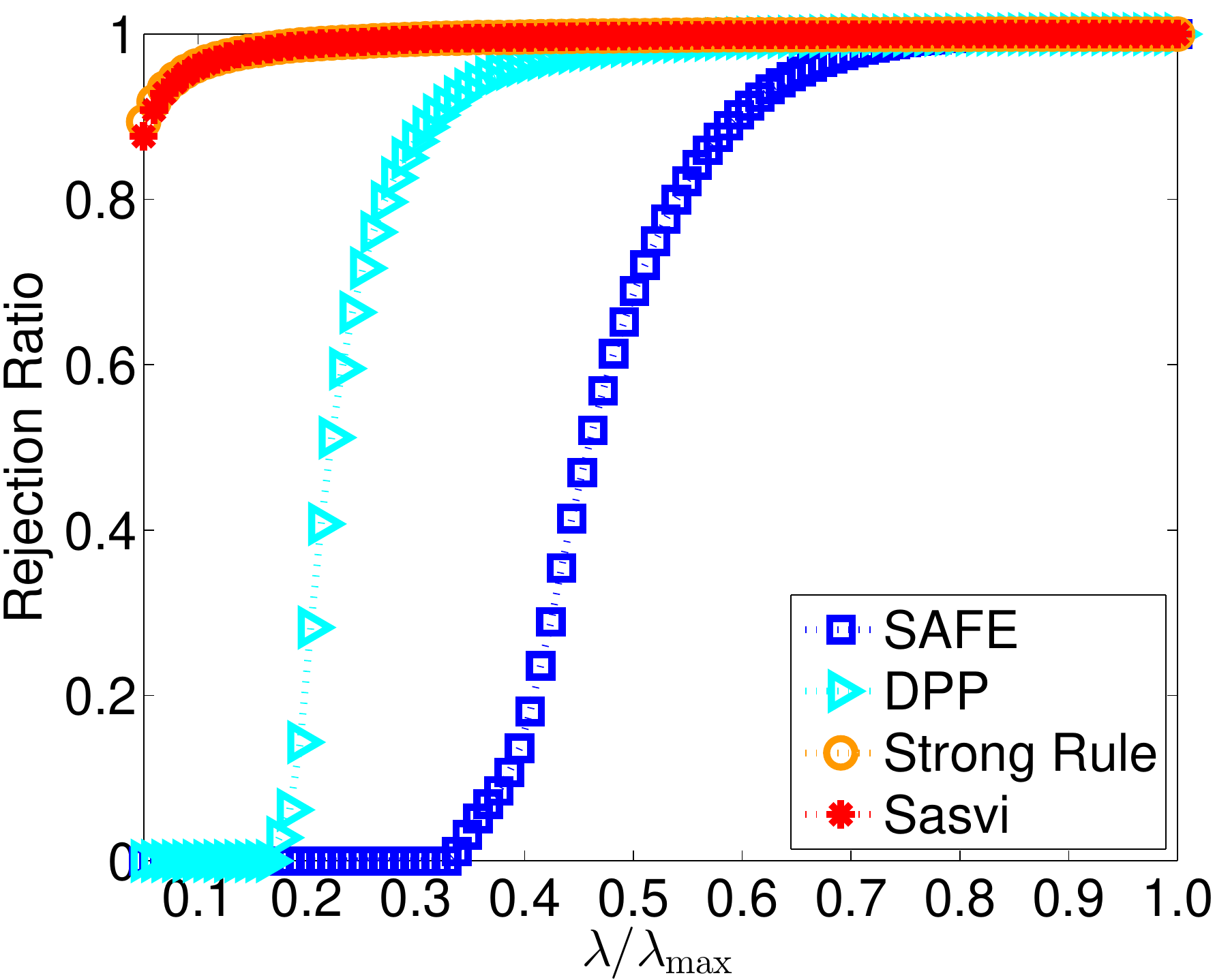}\\
 (Synthetic, $\bar p=5000$)\\
\caption{The rejectioin ratios---the ratios of the number features screened out by SAFE, DPP, the strong rule and Sasvi on synthetic and real data sets.}
\label{fig:lasso_synthetic}
\vspace{-0.1in}
\end{figure}

\noindent  {\bf Synthetic Data Set }  We follow~\cite{Bondell2008,Zou2005,Tibshirani:Lasso:1996} in simulating the data as follows:
\begin{equation}\label{eqn:regression_model}
  \mathbf y=X \bm \beta^*+\sigma \bm \epsilon, \quad \bm \epsilon \sim N(0,1),
\end{equation}
where $X$ has $250 \times 10000$ entries.
Similar to~\cite{Bondell2008,Zou2005,Tibshirani:Lasso:1996}, we set the pairwise correlation between the $i$-th feature and the $j$-th feature
to $0.5^{|i-j|}$ and draw $X$ from a Gaussian distribution. In constructing the ground truth $\bm \beta^*$,
we set the number of non-zero components to $\bar{p}$ and randomly assign the values from a uniform $[-1,1]$ distribution.
We set $\sigma=0.1$ and generate the response vector $\mathbf y \in  \mathbb{R}^{250}$ using Eq.~\eqref{eqn:regression_model}. For the 
value of $\bar{p}$, we try 100, 1000, and 5000.

\noindent  {\bf PIE Face Image Data Set } The PIE face image data set used in this experiment~\footnote{\url{http://www.cad.zju.edu.cn/home/dengcai/Data/FaceData.html}} contains $11554$ gray face images of $68$ people, taken under different poses,  illumination conditions and expressions. Each of the images has $32\times 32$ pixels. To use the regression model in Eq.~\eqref{eqn:regression_model}, we first randomly pick up an image as the response 
$\mathbf y \in  \mathbb{R}^{1024}$, and then set the remaining images as the data matrix $X \in  \mathbb{R}^{1024\times 11553}$. 

\noindent {\bf MNIST Handwritten Digit Data Set} This data set contains grey images of scanned handwritten digits, including $60,000$ for training and $10,000$ for testing. The dimension of each image is $28\times 28$. 
To use the regression model in Eq.~\eqref{eqn:regression_model}, we first randomly select $5000$ images for each digit from the training set (and in total we have $50000$ images) and get a data matrix $X \in  \mathbb{R} ^{784\times 50000}$,
and then we randomly select an image from the testing set and treat it as the response vector $\mathbf y \in \mathbb{R} ^{784}$. 

\noindent {\bf Experimental Settings } For the Lasso solver, we make use of the SLEP package~\cite{Liu:2009:SLEP:manual}.
For a given generated data set ($X$ and $\mathbf y$), we run the solver with or without screening rules to solve the Lasso problems along a sequence of $100$ parameter values equally spaced on the 
$\lambda/\lambda_{max}$ scale from $0.05$ to $1.0$. The reported results are averaged over $100$ trials of randomly drawn $X$ and $\mathbf y$.

\begin{table}
\centering
\begin{tabular}{|l|c|c|c|c|c| }
  \hline
  \multirow{2}{*}{Method} &  \multicolumn{3}{c|}{Synthetic with $\bar{p}$}               & \multicolumn{2}{|c|}{Real}  \\
  \cline{2-6}
	                        &  100 & 1000 & 5000 & MINST  & PIE \\
	\hline
	solver                  & 88.55          & 101.00        & 101.55        & 2683.57 & 617.85 \\
	SAFE                    & 73.37          & 88.42         & 90.21         & 651.23  & 128.54 \\
	DPP                     & 44.00          & 49.57         & 50.15         & 328.47  & 79.84  \\
	Strong                  & 2.53           & 3.00          & 2.92          & 5.57    & 2.97   \\
	Sasvi                   & 2.49           & 2.77          & 2.76          & 5.02    & 1.90   \\
	\hline
\end{tabular}
\caption{Running time (in seconds) for solving the Lasso problems along a sequence of $100$ tuning parameter values equally spaced on the scale of ${\lambda}/{\lambda_{max}}$ from $0.05$ to $1$ by 
the solver~\cite{Liu:2009:SLEP:manual} without screening, and the solver combined with different screening methods.}
\label{table:lasso_synthetic_time}
\vspace{-0.2in}
\end{table}


\noindent {\bf Results }  Table~\ref{table:lasso_synthetic_time} reports the running time by different screening rules, and 
Figure~\ref{fig:lasso_synthetic} presents the corresponding rejection ratios---the ratios of the number features screened out by the screening approaches.
It can be observed that the propose Sasvi significantly outperforms the safe screening rules such as SAFE and DPP. The reason is that, Sasvi
is able to discard more inactive features as discussed in Section~\ref{s:comparison}. 
In addition, the rejection ratios of the strong rule and Sasvi are comparable, 
and both of them are more effective in discarding inactive features than SAFE and DPP. 
In terms of the speedup, Sasvi provides better performance than the strong rule. 
The reason is that the strong rule is a heuristic screening method, i.e., it may mistakenly discard active features which have nonzero components in the solution,
and thus the strong rule needs to check the KKT conditions to make correction if necessary to ensure the correctness of the result. 
In contrast, Sasvi does not need to check the KKT conditions or make correction since the discarded features are guaranteed to be absent from the resulting sparse representation.

%% file: appendix.tex
\section{Proof of Theorem~\ref{theorem:main:result1}}\label{appendix:nonnegative}

\input{appendix_nonnegative}

\section{Proof of Theorem~\ref{theorem:min:result}}\label{appendix:min:result}

\input{appendix_Min_result}

\section{Proof of Theorem~\ref{theorem:main:result2}}\label{appendix:main:upper:bound}

\input{appendix_upper_bound}

\section{Proof of Theorem~\ref{theorem:main:result3}}\label{apendix:sure}

\input{appendix_feature_sure}

%% file: appendix_nonnegative.tex
We begin with three technical lemmas.

\begin{lemma}\label{eq:lemma:nonnegative}
Let $\mathbf y \neq \mathbf 0$ and $0<\lambda_1 \le \|X^T \mathbf y\|_{\infty}$. We have
\begin{equation}\label{eq:case:2:inequality:1}
   \langle  \frac{\mathbf y}{\lambda_1} - \bm \theta_1^*,  \bm \theta_1^* \rangle \ge 0.
\end{equation}
\end{lemma}
\noindent \textbf{Proof} 
Since the Euclidean projection of $\frac{\mathbf y}{\lambda_1}$ onto $\{\bm \theta: \| X^T  \bm \theta\|_{\infty} \le 1 \}$ is $\bm \theta_1^*$,
it follows from Lemma~\ref{lemma:optimization} that
\begin{equation}
  \langle \bm \theta_1^* - \frac{\mathbf y}{\lambda_1}, \bm \theta - \bm \theta_1^* \rangle  \ge 0, \forall \bm \theta: \| X^T  \bm \theta\|_{\infty} \le 1.
\end{equation}
As $\mathbf 0 \in \{\bm \theta: \| X^T  \bm \theta\|_{\infty} \le 1 \}$, we have Eq.~\eqref{eq:case:2:inequality:1}. \hfill $\Box$

\begin{lemma}\label{eq:lemma:non:parallel}
Let $\mathbf y \neq \mathbf 0$ and $0<\lambda_1 \le \|X^T \mathbf y\|_{\infty}$.
If $\bm \theta_1^*$ parallels to $\mathbf y$ in that
it can be written as $\bm \theta_1^* = \gamma \mathbf y$ for some $\gamma$,
then $\gamma = \frac{1}{\|X^T\mathbf y\|_{\infty}}$.
\end{lemma}
\noindent \textbf{Proof }  
Since $\frac{\mathbf y}{\|X^T\mathbf y\|_{\infty}}$ satisfies the
condition in Eq.~\eqref{eq:optimality:condition:1}, we have
\begin{equation}
    \langle \gamma \mathbf y - \frac{\mathbf y}{\lambda_1}, \frac{\mathbf y}{\|X^T\mathbf y\|_{\infty}} -  \gamma \mathbf y \rangle
		=(\gamma - \frac{1}{\lambda_1})(\frac{1}{\|X^T\mathbf y\|_{\infty}}- \gamma) \|\mathbf y\|_2^2 \ge 0
\end{equation}
which leads to $\gamma \in [\frac{1}{\|X^T\mathbf y\|_{\infty}}, \frac{1}{\lambda_1}]$.
In addition, since $\|X^T \bm \theta_1^* \|_{\infty} \le 1$, we have $\gamma = \frac{1}{\|X^T\mathbf y\|_{\infty}}$.
This completes the proof.
\hfill $\Box$

\begin{lemma}\label{eq:lemma:positive}
 Let $\mathbf y \neq \mathbf 0$. If $0< \lambda_1 \le \|X^T \mathbf y\|_{\infty}$, we have
\begin{equation}\label{eq:case:2:inequality:2}
   \langle  \frac{\mathbf y}{\lambda_1} - \bm \theta_1^*,  \mathbf y \rangle \geq 0,
\end{equation}
where the equality holds if and only if $\lambda_1 = \|X^T \mathbf y\|_{\infty}$.
\end{lemma}
\noindent \textbf{Proof}   We have
\begin{equation} \label{eq:inequality:positive}
 \langle  \frac{\mathbf y}{\lambda_1} - \bm \theta_1^*,  \frac{ \mathbf y }{\lambda_1}   \rangle  
	- \langle  \frac{\mathbf y}{\lambda_1} - \bm \theta_1^*,   \bm \theta_1^* \rangle
	=
   \langle  \frac{\mathbf y}{\lambda_1} - \bm \theta_1^*,  \frac{ \mathbf y }{\lambda_1} - \bm \theta_1^* \rangle 
	\ge 0,
\end{equation} 
where the equality holds if and only if $\frac{ \mathbf y }{\lambda_1} = \bm \theta_1^*  $. 
Incorporating Eq.~\eqref{eq:case:2:inequality:1} in Lemma~\ref{eq:lemma:nonnegative} and Eq.~\eqref{eq:inequality:positive}, we have Eq.~\eqref{eq:case:2:inequality:2}.
The equality in Eq.~\eqref{eq:inequality:positive} holds if and only if $\frac{ \mathbf y }{\lambda_1} = \bm \theta_1^* $.
According to Lemma~\ref{eq:lemma:non:parallel},  if $\bm \theta_1^*=\frac{ \mathbf y }{\lambda_1}$,
then $\bm \theta_1^*=\frac{ \mathbf y }{\|X^T \mathbf y\|_{\infty}}$, which leads
to $ \lambda_1 =\|X^T \mathbf y\|_{\infty}$.  This ends the proof. \hfill $\Box$

Now, we are ready to prove Theorem~\ref{theorem:main:result1}. If follows from Eq.~\eqref{eq:a:b:equation} and Eq.~\eqref{eq:case:2:inequality:2}
\begin{equation}\label{eq:b:a:inner:product}
   \langle \mathbf b, \mathbf a \rangle = 
	  (\frac{1}{\lambda_2} - \frac{1}{\lambda_1} ) \langle  \frac{\mathbf y}{\lambda_1} - \bm \theta_1^* ,  \mathbf y \rangle + \|\frac{\mathbf y}{\lambda_1} - \bm \theta_1^*\|_2^2
\end{equation}
\begin{equation}\label{eq:b:b:inner:product}
   \|\mathbf b\|_2^2 = 
	  \|(\frac{\mathbf y}{\lambda_2} - \frac{\mathbf y}{\lambda_1} ) \|_2^2+ 2 (\frac{1}{\lambda_2} - \frac{1}{\lambda_1} ) \langle  \frac{\mathbf y}{\lambda_1} - \bm \theta_1^* ,  \mathbf y \rangle + \|\frac{\mathbf y}{\lambda_1} - \bm \theta_1^*\|_2^2
		\ge 0.
\end{equation}
It follows from Lemma~\ref{eq:lemma:positive} that
1) $\langle \mathbf b, \mathbf a \rangle \ge 0$ and the equality holds if and only if $\frac{ \mathbf y }{\lambda_1} = \bm \theta_1^* $,
and 2) $\|\mathbf b\|_2^2 >0$, which leads to $\mathbf b \neq \mathbf 0$.
According to Lemma~\ref{eq:lemma:non:parallel}, if $\bm \theta_1^*$ parallels to $\mathbf y$, then
$\bm \theta_1^* = \frac{\mathbf y} {\|X^T \mathbf y\|_{\infty}}$.
Therefore, if $ 0< \lambda_1 < \|X^T \mathbf y\|_{\infty}$, then
$\langle \mathbf b, \mathbf a \rangle >0$ and $\mathbf a \neq 0$. \hfill $\Box$

%% file: appendix_Min_result.tex
If $\lambda_1=\|X^T \mathbf y\|_{\infty}$, the primal and dual optimals can be 
analytically computed as: $\bm \beta_1^* = \mathbf 0$ and $\bm \theta_1^*= \frac{\mathbf y} {\| X^T  \bm \theta\|_{\infty} }$.
Thus, we have $\mathbf a= \mathbf 0$. 
It is easy to get that $\mathbf r= -\frac{\mathbf x \|\mathbf b\|_2}{\|\mathbf x\|_2}$ minimizes Eq.~\eqref{eq:single:problem:study} with the minimum 
function value being 
\begin{equation}\label{eq:a:zero}
   \langle \mathbf x, \mathbf r \rangle = -\|\mathbf x\|_2\|\mathbf b\|_2.
\end{equation}

In our following discussion, we focus on the case $0< \lambda_1 <\|X^T \mathbf y\|_{\infty}$ and we have $\mathbf a \neq \mathbf 0$ according to Theorem~\ref{theorem:main:result1}.

The Lagrangian of Eq.~\eqref{eq:single:problem:study} can be written as
\begin{equation}\label{eq:lagrangian:problem:study}
 L(\mathbf r, \alpha, \beta)= \langle \mathbf x, \mathbf r \rangle + \alpha \langle \mathbf a, \mathbf r + \mathbf b \rangle + \frac{\beta}{2} (\|\mathbf r \|_2^2 - \|\mathbf b\|_2^2),
\end{equation}
where $\alpha, \beta \ge 0$ are introduced for the two inequalities, respectively.
It is clear that the minimal value of Eq.~\eqref{eq:single:problem:study} is lower bounded (the minimum is no less than $-\|\mathbf b \|_2 \|\mathbf x\|_2$ by only considering the constraint $\|\mathbf r \|_2^2 \le \|\mathbf b\|_2^2$).
Therefore, the optimal dual variable $\beta$ is always positive; otherwise, minimizing Eq.~\eqref{eq:lagrangian:problem:study} with regard to $\mathbf r$ achieves $-\infty$.

Setting the derivative with regard to $\mathbf r$ to zero, we have
\begin{equation}\label{eq:r:relationship}
  \mathbf r= \frac{- \mathbf x - \alpha \mathbf a}{\beta} .
\end{equation}

Plugging Eq.~\eqref{eq:r:relationship} into Eq.~\eqref{eq:lagrangian:problem:study}, we obtain the dual problem of Eq.~\eqref{eq:single:problem:study} as:
\begin{equation} \label{eq:additional:notations}
\begin{aligned}
  \max_{\alpha, \beta} & \quad  \alpha \langle \mathbf a, \mathbf b \rangle  - \frac{1}{2 \beta} \| \mathbf x + \alpha \mathbf a\|_2^2 - \frac{\beta}{2 }\|\mathbf b\|_2^2\\
	\mbox{subject to} & \quad  \alpha \ge 0, \beta \ge 0.
\end{aligned}
\end{equation}

For a given $\beta$, we have
\begin{equation}
\alpha =\max \left( \frac{\beta \langle \mathbf a, \mathbf b \rangle - \langle \mathbf x , \mathbf a \rangle}{\|\mathbf a\|_2^2}, 0 \right).
\end{equation}

We consider two cases. In the first case, we assume that $\alpha=0$. We have
\begin{equation}\label{eq:beta:less:than}
  \mathbf r= \frac{- \mathbf x  }{\beta},  \beta \le \frac{\langle \mathbf x , \mathbf a \rangle} {\langle \mathbf a, \mathbf b \rangle }.
\end{equation}
By using the complementary slackness condition (note that the optimal $\beta$ does not equal to zero), we have
\begin{equation}
   \|\mathbf r\|_2 = \left \| \frac{- \mathbf x  }{\beta} \right \|_2=\|\mathbf b\|_2.
\end{equation}
Thus, we have
\begin{equation}\label{eq:beta:equal}
  \beta =\frac{\|\mathbf x\|_2}{\|\mathbf b \|_2} .
\end{equation}
Incorporating Eq.~\eqref{eq:beta:less:than} and Eq.~\eqref{eq:beta:equal}, we have
\begin{equation}\label{eq:condition:case:1:condition}
  \frac{ \langle \mathbf b, \mathbf a \rangle }{ \|\mathbf b \|_2  \|\mathbf a \|_2 } \le \frac{\langle \mathbf x , \mathbf a \rangle} { \|\mathbf x\|_2  \|\mathbf a \|_2 },
\end{equation}
so that the angle between $\mathbf a$ and $\mathbf b$ is equal to or larger than 
the angle between $\mathbf x$ and $\mathbf a$. Note that $\langle \mathbf b, \mathbf a \rangle \ge 0$ according to Theorem~\ref{theorem:main:result1}.
In Figure~\ref{fig:explain_theorem}, EX$_2$ and EX$_3$ illustrate
the case that $\mathbf x$ satisfies Eq.~\eqref{eq:condition:case:1:condition}, 
while EX$_1$ and EX$_4$ show the opposite cases.
In addition, we have
\begin{equation}\label{eq:condition:case:1}
   \langle \mathbf x, \mathbf r \rangle = - \|\mathbf x\|_2 \|\mathbf b \|_2.
\end{equation}

In the second case, Eq.~\eqref{eq:condition:case:1:condition} does not hold. We have
\begin{equation}\label{eq:alpha:non:zero}
   \alpha = \frac{\beta \langle \mathbf a, \mathbf b \rangle - \langle \mathbf x , \mathbf a \rangle}{\|\mathbf a\|_2^2}.
\end{equation}
Plugging Eq.~\eqref{eq:alpha:non:zero} into Eq.~\eqref{eq:r:relationship}, we have
\begin{equation}\label{eq:r:case:2}
  \mathbf r= - \frac{ \mathbf x \|\mathbf a\|_2^2 +   \beta \langle \mathbf a, \mathbf b \rangle \mathbf a- \langle \mathbf x , \mathbf a \rangle \mathbf a }{\beta \|\mathbf a\|_2^2}
\end{equation}
Since $\|\mathbf r\|_2^2=\|\mathbf b\|_2^2$, we have
\begin{equation}\label{eq:beta:2:solution}
  \beta=  \sqrt{\frac{ \|\mathbf x\|_2^2  \|\mathbf a\|_2^2- \langle \mathbf x, \mathbf a \rangle^2   }{  \|\mathbf b\|_2^2  \|\mathbf a\|_2^2- \langle \mathbf b, \mathbf a \rangle^2  } } =
	\frac{ \|\mathbf x^{\bot}\|_2 }{ \sqrt{ \|\mathbf b\|_2^2 - \frac{\langle \mathbf b, \mathbf a \rangle^2}{  \|\mathbf a\|_2^2}  } },
\end{equation}
where we have used Eq.~\eqref{eq:x:bot:def} to get the second equality.
In addition, we have
\begin{equation}\label{eq:condition:case:2}
   \langle \mathbf x, \mathbf r \rangle  =  
	-  \|\mathbf x^{\bot}\|_2 \sqrt{\|\mathbf b\|_2^2 - \frac{ \langle \mathbf b, \mathbf a \rangle ^2}{ \|\mathbf a\|_2^2} } 																			 
	-\frac{ \langle \mathbf a, \mathbf b \rangle \langle \mathbf x, \mathbf a \rangle} { \|\mathbf a\|_2^2}.
\end{equation}

In summary, Eq.~\eqref{eq:single:problem:study} equals to $-\|\mathbf x\|_2\|\mathbf b\|_2$, if $\frac{ \langle \mathbf b, \mathbf a \rangle }{ \|\mathbf b \|_2 } \le \frac{\langle \mathbf x , \mathbf a \rangle} { \|\mathbf x\|_2 }$, and $-  \|\mathbf x^{\bot}\|_2 \sqrt{\|\mathbf b\|_2^2 - \frac{ \langle \mathbf b, \mathbf a \rangle ^2}{ \|\mathbf a\|_2^2} } 																			 
	-\frac{ \langle \mathbf a, \mathbf b \rangle \langle \mathbf x, \mathbf a \rangle} { \|\mathbf a\|_2^2}$ otherwise. 
This ends the proof of this theorem. \hfill $\Box$

%% file: appendix_upper_bound.tex
We prove the four cases one by one as follows.

\noindent \textbf{Case 1 }  If $\mathbf a \neq \mathbf 0$ and $\frac{ \langle \mathbf b, \mathbf a \rangle }{\|\mathbf b \|_2   } > \frac{ | \langle \mathbf x_j , \mathbf a \rangle |} { \|\mathbf x_j\|_2  }$,
i.e., Eq.~\eqref{eq:condition:case:1:condition} does not hold with $\mathbf x = \pm \mathbf x_j$. We have
\begin{equation}\label{eq:proof:theorem:positive}
					   \begin{aligned}
					   u_j^+(\lambda_2) 
						& = \max_{\bm \theta: \langle \bm \theta_1^* - \frac{\mathbf y}{\lambda_1}, \bm \theta - \bm \theta_1^* \rangle \ge 0,
\langle \bm \theta - \frac{\mathbf y}{\lambda_2}, \bm \theta_1^* - \bm \theta \rangle \ge 0
}  \langle \mathbf x_j, \bm \theta  \rangle  \\
            & =  	\frac{ 1}{2} \max_{\mathbf r: \langle    \mathbf a, \mathbf r + \mathbf b \rangle \le 0,      \|\mathbf r \|_2^2 \le \|\mathbf b \|_2^2}   						
					 \left[ \langle \mathbf x_j,    \bm \theta_1^* + \frac{\mathbf y}{\lambda_2}   \rangle +  \langle \mathbf x_j, \mathbf r  \rangle			\right]		\\
						& = \frac{ 1}{2} \left[ \langle \mathbf x_j,     \bm \theta_1^* + \frac{\mathbf y}{\lambda_2}   \rangle  +	\max_{\mathbf r: \langle    \mathbf a, \mathbf r + \mathbf b \rangle \le 0,      \|\mathbf r \|_2^2 \le \|\mathbf b \|_2^2}    \langle \mathbf x_j, \mathbf r  \rangle				\right]	\\
						& =  \frac{ 1}{2} \left[  \langle \mathbf x_j,     \bm \theta_1^* + \frac{\mathbf y}{\lambda_2}  \rangle  -	 \min_{\mathbf r: \langle    \mathbf a, \mathbf r + \mathbf b \rangle \le 0,      \|\mathbf r \|_2^2 \le \|\mathbf b \|_2^2}    \langle - \mathbf x_j, \mathbf r  \rangle		\right]			\\
						& =  \frac{ 1}{2} \left[  \langle \mathbf x_j,     2 \bm \theta_1^*   + (\frac{\mathbf y}{\lambda_1}  - \bm \theta_1^* ) +  (\frac{\mathbf y}{\lambda_2}  - \frac{\mathbf y}{\lambda_1} )  \rangle \right]	  \\						
						&    \quad  +  \frac{ 1}{2} \left[
						   \|-\mathbf x_j^{\bot}\|_2 \sqrt{ 
	                                          \|\mathbf b\|_2^2 - \frac{ \langle \mathbf b, \mathbf a \rangle ^2}{ \|\mathbf a\|_2^2} }  
																							+\frac{ \langle \mathbf a, \mathbf b \rangle \langle -\mathbf x_j, \mathbf a \rangle} { \|\mathbf a\|_2^2} \right]		 \\						
						& =   \langle \mathbf x_j, \bm \theta_1^* \rangle + \frac{\frac{1}{\lambda_2} - \frac{1}{\lambda_1}}{2} [ \langle \mathbf x_j , \mathbf y \rangle
- 
 \frac{\langle  \mathbf a  ,  \mathbf y  \rangle}{\|  \mathbf a  \|_2^2}     \langle \mathbf x_j,  \mathbf a \rangle ]										\\
						&   \quad  + 
						\frac{\frac{1}{\lambda_2} - \frac{1}{\lambda_1}}{2}  \|\mathbf x_j^{\bot}\|_2\sqrt{\|\mathbf y\|_2^2  - \frac{\langle \mathbf y, \mathbf a\rangle ^2}{\|\mathbf a\|_2^2} }.
						\end{aligned}
					\end{equation}
The second equality plugs in the notations in Eq.~\eqref{eq:a:b:equation}.
The fifth equality utilizes Eq.~\eqref{eq:condition:case:2} which is the result for the case $\frac{ \langle \mathbf b, \mathbf a \rangle }{\|\mathbf b \|_2   } 
> \frac{ | \langle \mathbf x_j , \mathbf a \rangle |} { \|\mathbf x_j\|_2  }   \ge \frac{   \langle -\mathbf x_j , \mathbf a \rangle } { \|\mathbf x_j\|_2  }$ by setting $\mathbf x = - \mathbf x_j$.
To get the last equality, we utlize the following two equalities
\begin{equation}\label{eq:b:a:y:relationship:1}
					   \|\mathbf b\|_2^2  - \frac{\langle \mathbf b, \mathbf a\rangle ^2}{\|\mathbf a\|_2^2} = ( \frac{1}{\lambda_2} - \frac{1}{\lambda_1})^2
						 (\|\mathbf y\|_2^2  - \frac{\langle \mathbf y, \mathbf a\rangle ^2}{\|\mathbf a\|_2^2})
\end{equation}
and 
\begin{equation}
					   \frac{ \langle \mathbf a, \mathbf b \rangle \langle \mathbf x_j, \mathbf a \rangle} { \|\mathbf a\|_2^2} = \langle \mathbf x_j, \mathbf a \rangle  
						(1 + \frac{ \langle \mathbf a, \mathbf y  \rangle  (\frac{1}{\lambda_2}- \frac{1}{\lambda_1})  }   { \|\mathbf a\|_2^2}),
\end{equation}
which can be derived from Eq.~\eqref{eq:a:b:equation}. 
It follows from Eq.~\eqref{eq:x:j:bot} and Eq.~\eqref{eq:y:bot} that
\begin{equation}\label{eq:x:bot:relationship}
    \|\mathbf x_j^{\bot} \|_2^2 = \|\mathbf x_j\|_2^2 - \frac{ \langle \mathbf x_j, \mathbf a \rangle ^2}{  \|\mathbf a\|_2^2},
\end{equation}
\begin{equation}\label{eq:y:bot:relationship}
    \|\mathbf y^{\bot} \|_2^2 = \|\mathbf y\|_2^2  - \frac{\langle \mathbf y, \mathbf a\rangle ^2}{\|\mathbf a\|_2^2},
\end{equation}
\begin{equation}\label{eq:x::ybot:relationship}
   \langle \mathbf x_j^{\bot},\mathbf y^{\bot} \rangle =  \langle \mathbf x_j , \mathbf y \rangle - 
 \frac{\langle  \mathbf a  ,  \mathbf y  \rangle}{\|  \mathbf a  \|_2^2}     \langle \mathbf x_j,  \mathbf a \rangle.
\end{equation}
Incorporating Eq.~\eqref{eq:proof:theorem:positive}, and Eqs.~\eqref{eq:y:bot:relationship}-\eqref{eq:x::ybot:relationship},
we have Eq.~\eqref{eq:bound:little:lambda:positive}. Following a similar derivation, we have
\begin{equation}\label{eq:proof:theorem:negative}
					   \begin{aligned}
					  u_j^-(\lambda_2) 
						& = \max_{\bm \theta: \langle \bm \theta_1^* - \frac{\mathbf y}{\lambda_1}, \bm \theta - \bm \theta_1^* \rangle \ge 0,
\langle \bm \theta - \frac{\mathbf y}{\lambda_2}, \bm \theta_1^* - \bm \theta \rangle \ge 0
}  \langle -\mathbf x_j, \bm \theta  \rangle  \\
            & =  \frac{1}{2} 	\max_{\mathbf r: \langle    \mathbf a, \mathbf r + \mathbf b \rangle \le 0,      \|\mathbf r \|_2^2 \le \|\mathbf b \|_2^2}  \left[  \langle -\mathbf x_j,     \bm \theta_1^* + \frac{\mathbf y}{\lambda_2}  \rangle 
						     +  \langle -\mathbf x_j, \mathbf r  \rangle				\right]	\\
						& =   \frac{1}{2} \left[	 \langle -\mathbf x_j,     \bm \theta_1^* + \frac{\mathbf y}{\lambda_2}    \rangle  +	\max_{\mathbf r: \langle    \mathbf a, \mathbf r + \mathbf b \rangle \le 0,      \|\mathbf r \|_2^2 \le \|\mathbf b \|_2^2}    \langle - \mathbf x_j, \mathbf r  \rangle			\right]		\\
						& =  \frac{1}{2} \left[ \langle - \mathbf x_j,    \bm \theta_1^* + \frac{\mathbf y}{\lambda_2}  \rangle  -	\min_{\mathbf r: \langle    \mathbf a, \mathbf r + \mathbf b \rangle \le 0,      \|\mathbf r \|_2^2 \le \|\mathbf b \|_2^2}    \langle  \mathbf x_j, \mathbf r  \rangle			\right]		\\
						& = \frac{1}{2} \left[  	\langle - \mathbf x_j,  2 \bm \theta_1^*   + (\frac{\mathbf y}{\lambda_1}  - \bm \theta_1^* ) +  (\frac{\mathbf y}{\lambda_2}  - \frac{\mathbf y}{\lambda_1} )  \rangle \right] \\						
						&   \quad   + 
						 \frac{1}{2}  \left[ \|\mathbf x_j^{\bot}\|_2 \sqrt{\|\mathbf b\|_2^2- \frac{ \langle \mathbf b, \mathbf a \rangle ^2}{ \|\mathbf a\|_2^2} }  
																							+\frac{ \langle \mathbf a, \mathbf b \rangle \langle \mathbf x_j, \mathbf a \rangle} { \|\mathbf a\|_2^2} \right] \\						
						& =  - \langle \mathbf x_j, \bm \theta_1^* \rangle - \frac{\frac{1}{\lambda_2} - \frac{1}{\lambda_1}}{2} [ \langle \mathbf x_j , \mathbf y \rangle
- 
 \frac{\langle   \mathbf a ,  \mathbf y  \rangle}{\|  \mathbf a  \|_2^2}     \langle \mathbf x_j,  \mathbf a  \rangle ]										\\
						&   \quad    + 
						\frac{\frac{1}{\lambda_2} - \frac{1}{\lambda_1}}{2}  \|\mathbf x_j^{\bot}\|_2  \sqrt{
	                                           \|\mathbf y\|_2^2  - \frac{\langle \mathbf y, \mathbf a\rangle ^2}{\|\mathbf a\|_2^2} }.
						\end{aligned}
\end{equation}
The fifth equality utilizes Eq.~\eqref{eq:condition:case:2} which is the result for the case $\frac{ \langle \mathbf b, \mathbf a \rangle }{\|\mathbf b \|_2   } 
> \frac{ | \langle \mathbf x_j , \mathbf a \rangle |} { \|\mathbf x_j\|_2  }   \ge \frac{   \langle \mathbf x_j , \mathbf a \rangle } { \|\mathbf x_j\|_2  }$ by setting $\mathbf x =   \mathbf x_j$.	
The last equality can be obtained using the similar derivation getting the last equality of Eq.~\eqref{eq:proof:theorem:positive}.				
Incorporating Eqs.~\eqref{eq:y:bot:relationship}-\eqref{eq:proof:theorem:negative},
we have Eq.~\eqref{eq:bound:little:lambda:negative}.
					
\noindent \textbf{Case 2 } 
If $\frac{ \langle \mathbf b, \mathbf a \rangle }{\|\mathbf b \|_2   } \le \frac{  \langle \mathbf x_j , \mathbf a \rangle  } { \|\mathbf x_j\|_2  }$ and $\langle \mathbf x_j , \mathbf a \rangle >0$,
we have $\frac{ \langle \mathbf b, \mathbf a \rangle }{\|\mathbf b \|_2   } > \frac{  \langle - \mathbf x_j , \mathbf a \rangle  } { \|\mathbf x_j\|_2  }$ since $\langle \mathbf b, \mathbf a \rangle \ge 0$ according to Theorem~\ref{theorem:main:result1}. Thus, Eq.~\eqref{eq:condition:case:1:condition} does not hold with $\mathbf x = - \mathbf x_j$, and we can 
get Eq.~\eqref{eq:proof:theorem:positive}, or equivalently Eq.~\eqref{eq:bound:little:lambda:positive}. 
In addition, Eq.~\eqref{eq:condition:case:1:condition} holds with $\mathbf x =   \mathbf x_j$, and we have
\begin{equation}\label{eq:proof:theorem:negative:2}
					   \begin{aligned}
					  u_j^-(\lambda_2) 
						& = \max_{\bm \theta: \langle \bm \theta_1^* - \frac{\mathbf y}{\lambda_1}, \bm \theta - \bm \theta_1^* \rangle \ge 0,
\langle \bm \theta - \frac{\mathbf y}{\lambda_2}, \bm \theta_1^* - \bm \theta \rangle \ge 0
}   \langle -\mathbf x_j, \bm \theta  \rangle  \\
            & =  	\max_{\mathbf r: \langle    \mathbf a, \mathbf r + \mathbf b \rangle \le 0,      \|\mathbf r \|_2^2 \le \|\mathbf b \|_2^2}   
						 \left[  \langle -\mathbf x_j,   
						  \frac{\bm \theta_1^* + \frac{\mathbf y}{\lambda_2} }{2}  \rangle +  \frac{1}{2} \langle -\mathbf x_j, \mathbf r  \rangle		\right]			\\
						& =  \langle -\mathbf x_j,    \frac{\bm \theta_1^* + \frac{\mathbf y}{\lambda_2} }{2}  \rangle  
						      + \frac{1}{2} \max_{\mathbf r: \langle    \mathbf a, \mathbf r + \mathbf b \rangle \le 0,      \|\mathbf r \|_2^2 \le \|\mathbf b \|_2^2}    \langle - \mathbf x_j, \mathbf r  \rangle					\\
						& =  \langle - \mathbf x_j,    \frac{\bm \theta_1^* + \frac{\mathbf y}{\lambda_2} }{2}  \rangle  
						      -	\frac{1}{2} \min_{\mathbf r: \langle    \mathbf a, \mathbf r + \mathbf b \rangle \le 0,      \|\mathbf r \|_2^2 \le \|\mathbf b \|_2^2}    \langle  \mathbf x_j, \mathbf r  \rangle					\\
						& =  	\langle - \mathbf x_j, \bm \theta_1^*  + \frac{1}{2} ( \frac{\mathbf y}{\lambda_2} - \bm \theta_1^*) \rangle  + \frac{1}{2}  \|\mathbf x_j\|_2 \|\mathbf b\|_2 \\						
						& =     - \langle \mathbf x_j, \bm \theta_1^* \rangle +
						 \frac{1}{2} [\|\mathbf x_j\|_2 \|\mathbf b\|_2 - \langle \mathbf x_j, \mathbf b \rangle ] .
						\end{aligned}
\end{equation}
To get the fifth equality, we utilize Eq.~\eqref{eq:condition:case:1} with $\mathbf x =\mathbf x_j$.  
Therefore, we have Eq.~\eqref{eq:bound:little:lambda:negative:2}.

\noindent \textbf{Case 3 } 
If $\frac{ \langle \mathbf b, \mathbf a \rangle }{\|\mathbf b \|_2   } \le \frac{  -\langle \mathbf x_j , \mathbf a \rangle  } { \|\mathbf x_j\|_2  }$ and $\langle \mathbf x_j , \mathbf a \rangle  <0$,
Eq.~\eqref{eq:condition:case:1:condition} holds with $\mathbf x = -  \mathbf x_j$, and we have
\begin{equation}\label{eq:proof:theorem:positive:2}
					   \begin{aligned}
	  u_j^+(\lambda_2) 
						& = \max_{\bm \theta: \langle \bm \theta_1^* - \frac{\mathbf y}{\lambda_1}, \bm \theta - \bm \theta_1^* \rangle \ge 0,
\langle \bm \theta - \frac{\mathbf y}{\lambda_2}, \bm \theta_1^* - \bm \theta \rangle \ge 0
} 
 \langle \mathbf x_j, \bm \theta  \rangle  \\
            & =  	\max_{\mathbf r: \langle    \mathbf a, \mathbf r + \mathbf b \rangle \le 0,      \|\mathbf r \|_2^2 \le \|\mathbf b \|_2^2}  
						 \left[ \langle \mathbf x_j,    \frac{\bm \theta_1^* + \frac{\mathbf y}{\lambda_2} }{2}  \rangle +  \frac{1}{2} \langle \mathbf x_j, \mathbf r  \rangle				\right]	\\
						& =  \langle \mathbf x_j,    \frac{\bm \theta_1^* + \frac{\mathbf y}{\lambda_2} }{2}  \rangle  +	\frac{1}{2}\max_{\mathbf r: \langle    \mathbf a, \mathbf r + \mathbf b \rangle \le 0,      \|\mathbf r \|_2^2 \le \|\mathbf b \|_2^2}    \langle \mathbf x_j, \mathbf r  \rangle					\\
						& =  \langle \mathbf x_j,    \frac{\bm \theta_1^* + \frac{\mathbf y}{\lambda_2} }{2}  \rangle  -	\frac{1}{2} \min_{\mathbf r: \langle    \mathbf a, \mathbf r + \mathbf b \rangle \le 0,      \|\mathbf r \|_2^2 \le \|\mathbf b \|_2^2}    \langle - \mathbf x_j, \mathbf r  \rangle					\\
					  & =  	\langle  \mathbf x_j, \bm \theta_1^*  + \frac{1}{2} ( \frac{\mathbf y}{\lambda_2} - \bm \theta_1^*)  \rangle  + \frac{1}{2} \|-\mathbf x_j\|_2 \|\mathbf b\|_2 \\						
						& =      \langle \mathbf x_j, \bm \theta_1^* \rangle + \frac{1}{2} [\|\mathbf x_j\|_2 \|\mathbf b\|_2 + \langle \mathbf x_j, \mathbf b \rangle ] ,
						\end{aligned}
\end{equation}
where the fifth equality utilizes Eq.~\eqref{eq:condition:case:1} with $\mathbf x =  - \mathbf x_j$.  Therefore, we have Eq.~\eqref{eq:bound:little:lambda:positive:2}.
In addition, we have $\frac{ \langle \mathbf b, \mathbf a \rangle }{\|\mathbf b \|_2   } > \frac{  \langle  \mathbf x_j , \mathbf a \rangle  } { \|\mathbf x_j\|_2  }$ since $\langle \mathbf b, \mathbf a \rangle \ge 0$ according to Theorem~\ref{theorem:main:result1} and  $\langle \mathbf x_j , \mathbf a \rangle  <0$. Thus, Eq.~\eqref{eq:condition:case:1:condition} does not hold with $\mathbf x = \mathbf x_j$, and we can 
get Eq.~\eqref{eq:proof:theorem:negative}, or equivalently Eq.~\eqref{eq:bound:little:lambda:negative}.

\noindent \textbf{Case 4 } 
If $ \mathbf a= \mathbf 0$, then we have $\lambda_1=\|X^T \mathbf y\|_{\infty}$ according to Theorem~\ref{theorem:main:result1}.
Therefore,
\begin{equation}\label{eq:proof:theorem:positive:3}
					   \begin{aligned}
					   u_j^+(\lambda_2) 
						& = \max_{\bm \theta: \langle \bm \theta_1^* - \frac{\mathbf y}{\lambda_1}, \bm \theta - \bm \theta_1^* \rangle \ge 0,
\langle \bm \theta - \frac{\mathbf y}{\lambda_2}, \bm \theta_1^* - \bm \theta \rangle \ge 0
}  \langle \mathbf x_j, \bm \theta  \rangle  \\
            & =  	\frac{ 1}{2} \max_{\mathbf r: \langle    \mathbf a, \mathbf r + \mathbf b \rangle \le 0,      \|\mathbf r \|_2^2 \le \|\mathbf b \|_2^2}   						
					 \left[ \langle \mathbf x_j,    \bm \theta_1^* + \frac{\mathbf y}{\lambda_2}   \rangle +  \langle \mathbf x_j, \mathbf r  \rangle			\right]		\\
						& = \frac{ 1}{2} \left[ \langle \mathbf x_j,     \bm \theta_1^* + \frac{\mathbf y}{\lambda_2}   \rangle  +	\max_{\mathbf r: \langle    \mathbf a, \mathbf r + \mathbf b \rangle \le 0,      \|\mathbf r \|_2^2 \le \|\mathbf b \|_2^2}    \langle \mathbf x_j, \mathbf r  \rangle				\right]	\\
						& =  \frac{ 1}{2} \left[  \langle \mathbf x_j,     \bm \theta_1^* + \frac{\mathbf y}{\lambda_2}  \rangle  -	 \min_{\mathbf r: \langle    \mathbf a, \mathbf r + \mathbf b \rangle \le 0,      \|\mathbf r \|_2^2 \le \|\mathbf b \|_2^2}    \langle - \mathbf x_j, \mathbf r  \rangle		\right]			\\
						& =  \frac{ 1}{2} \left[  \langle \mathbf x_j,     2 \bm \theta_1^*   + (\frac{\mathbf y}{\lambda_2}  - \bm \theta_1^* )   \rangle \right]	  + \frac{1}{2} \|-\mathbf x_j\|_2\|\mathbf b\|_2
	\end{aligned}
\end{equation}
To get the last equality, we utilize Eq.~\eqref{eq:a:zero} with $\mathbf x=-\mathbf x_j$. Therefore, we have Eq.~\eqref{eq:proof:theorem:positive:2}.
Similarly,
\begin{equation}\label{eq:proof:theorem:negative:3}
					   \begin{aligned}
					  u_j^-(\lambda_2) 
						& = \max_{\bm \theta: \langle \bm \theta_1^* - \frac{\mathbf y}{\lambda_1}, \bm \theta - \bm \theta_1^* \rangle \ge 0,
\langle \bm \theta - \frac{\mathbf y}{\lambda_2}, \bm \theta_1^* - \bm \theta \rangle \ge 0
}   \langle -\mathbf x_j, \bm \theta  \rangle  \\
            & =  	\max_{\mathbf r: \langle    \mathbf a, \mathbf r + \mathbf b \rangle \le 0,      \|\mathbf r \|_2^2 \le \|\mathbf b \|_2^2}   
						 \left[  \langle -\mathbf x_j,   
						  \frac{\bm \theta_1^* + \frac{\mathbf y}{\lambda_2} }{2}  \rangle +  \frac{1}{2} \langle -\mathbf x_j, \mathbf r  \rangle		\right]			\\
						& =  \langle -\mathbf x_j,    \frac{\bm \theta_1^* + \frac{\mathbf y}{\lambda_2} }{2}  \rangle  
						      + \frac{1}{2} \max_{\mathbf r: \langle    \mathbf a, \mathbf r + \mathbf b \rangle \le 0,      \|\mathbf r \|_2^2 \le \|\mathbf b \|_2^2}    \langle - \mathbf x_j, \mathbf r  \rangle					\\
						& =  \langle - \mathbf x_j,    \frac{\bm \theta_1^* + \frac{\mathbf y}{\lambda_2} }{2}  \rangle  
						      -	\frac{1}{2} \min_{\mathbf r: \langle    \mathbf a, \mathbf r + \mathbf b \rangle \le 0,      \|\mathbf r \|_2^2 \le \|\mathbf b \|_2^2}    \langle  \mathbf x_j, \mathbf r  \rangle					\\
						& =  	\langle - \mathbf x_j, \bm \theta_1^*  + \frac{1}{2} ( \frac{\mathbf y}{\lambda_2} - \bm \theta_1^*) \rangle  + \frac{1}{2}  \|\mathbf x_j\|_2 \|\mathbf b\|_2 
						\end{aligned}
\end{equation}
To get the last equality, we utilize Eq.~\eqref{eq:a:zero} with $\mathbf x=\mathbf x_j$.  Therefore, we have Eq.~\eqref{eq:proof:theorem:negative:2}. 

This ends the proof of this theorem. \hfill $\Box$

%% file: appendix_feature_sure.tex
We begin with a technical lemma. For a geometrical illustration of this lemma, please refer to the first plot of Figure~\ref{fig:sureRemove}.

\begin{lemma}\label{eq:lemma:monotone}
Let $\mathbf y \neq 0$, and $\|X^T \mathbf y\|_{\infty} > \lambda_1 > \lambda >0$.
Suppose that $\bm \theta_1^* \neq \frac{\mathbf y}{\|X^T\mathbf y\|_{\infty}}$.
For the two auxiliary functions defined in Eq.~\eqref{eq:aux:f} and Eq.~\eqref{eq:aux:g},
$f(\lambda)$ is strictly increasing with regard to $\lambda$ in $(0,  \lambda_1 ]$. 
$g(\lambda)$ is strictly decreasing with regard to $\lambda$ in $(0,  \lambda_1 ]$. 
\end{lemma}
\noindent \textbf{Proof }  Denote $\gamma =\frac{1}{\lambda} - \frac{1}{\lambda_1}$. 
We can rewrite $f(\lambda)$ as
\begin{equation}
  h(\gamma)= \frac{ \langle \mathbf a + \gamma \mathbf y, \mathbf a \rangle }{\|   \mathbf a + \gamma \mathbf y \|_2  }.
\end{equation}
The derivative of $h(\gamma)$ with regard to $\gamma$ can be computed as
\begin{equation}\label{eq:h:gamma:nonnegative}
  h'(\gamma)= \frac{ \gamma (  \langle \mathbf a, \mathbf y \rangle^2 - \|\mathbf y\|_2^2 \|\mathbf a\|_2^2  )}{\|\mathbf a + \gamma \mathbf y \|_2^3} \le 0
\end{equation}
For any $\gamma >0$, $h'(\gamma)=0$ if and only if $\mathbf a$  parallels to $\mathbf y$.
It follows the definition of $\mathbf a$ in Eq.~\eqref{eq:a:b:equation} that, 
if $\mathbf a$  parallels to $\mathbf y$, then $\bm \theta_1^*$ parallels $\mathbf y$.
According to Lemma~\ref{eq:lemma:non:parallel}, 
we have $\bm \theta_1^* = \frac{\mathbf y}{\|X^T\mathbf y\|_{\infty}}$,
which contradicts to the assumption $\bm \theta_1^* \neq \frac{\mathbf y}{\|X^T\mathbf y\|_{\infty}}$.
Therefore, $h'(\gamma)>0$, $h(\gamma)$ is strictly decreasing $\forall \gamma >0$, and $f(\lambda)$ is strictly increasing with regard to $\lambda$ in $(0,  \lambda_1 ]$. 
Following a similar proof, we can show that $g(\lambda)$ is strictly decreasing with regard to $\lambda$ in $(0,  \lambda_1 ]$.
\hfill $\Box$

Now, are ready to prove Theorem~\ref{theorem:main:result3}. Firstly, we summarize the $u_j^+ (\lambda_2)$ and $u_j^- (\lambda_2)$ 
in unified equations.

Since $ \langle \mathbf x_j, \mathbf a \rangle \ge 0$, $ u_j^+ (\lambda_2)$ satisfies Eq.~\eqref{eq:bound:little:lambda:positive} if $\mathbf a \neq 0$,
and Eq.~\eqref{eq:bound:little:lambda:positive:2} otherwise. Thus, we have
\begin{equation}\label{eq:x:j:upper:bound}
   u_j^+ (\lambda_2) =\left\{
	\begin{aligned}	
	   &  \langle \mathbf x_j, \bm \theta_1^* \rangle +  \frac{\frac{1}{\lambda_2} - \frac{1}{\lambda_1}}{2} \left [ \|\mathbf x_j^{\bot}\|_2 \|\mathbf y^{\bot}\|_2  + \langle \mathbf x_j^{\bot} , \mathbf y ^{\bot} \rangle \right], 
		    &  \quad   \mathbf a \neq \mathbf 0  \\						
	    &\langle \mathbf x_j, \bm \theta_1^* \rangle +  \frac{1}{2} \left[ \|\mathbf x_j\|_2 \|\mathbf b\|_2 + \langle \mathbf x_j, \mathbf b \rangle \right],
			   & \quad  \mathbf a = \mathbf 0 \\
	\end{aligned}
	  \right.
\end{equation}

Since $ \langle \mathbf x_j, \mathbf a \rangle \ge 0$, $ u_j^- (\lambda_2)$ satisfies Eq.~\eqref{eq:bound:little:lambda:negative:2} if $\frac{ \langle \mathbf b, \mathbf a \rangle }{\|\mathbf b \|_2   } \le \frac{  \langle \mathbf x_j , \mathbf a \rangle } { \|\mathbf x_j\|_2  }  $,
and Eq.~\eqref{eq:bound:little:lambda:negative} otherwise. Thus, we have
\begin{equation}\label{eq:minus:x:j:upper:bound}
   u_j^- (\lambda_2) =
	 \left\{
	\begin{aligned}
	   & -  \langle \mathbf x_j, \bm \theta_1^* \rangle + \frac{1}{2} \left[ \|\mathbf x_j\|_2 \|\mathbf b\|_2- \langle \mathbf x_j, \mathbf b \rangle \right],  &  \quad
		    \frac{ \langle \mathbf b, \mathbf a \rangle }{\|\mathbf b \|_2   } \le \frac{  \langle \mathbf x_j , \mathbf a \rangle } { \|\mathbf x_j\|_2  }   \\		
	   & - \langle \mathbf x_j, \bm \theta_1^* \rangle + \frac{\frac{1}{\lambda_2} - \frac{1}{\lambda_1}}{2} \left [ \|\mathbf x_j^{\bot}\|_2 \|\mathbf y^{\bot}\|_2  - \langle \mathbf x_j^{\bot} , \mathbf y ^{\bot} \rangle \right], & 
	      \quad \frac{ \langle \mathbf b, \mathbf a \rangle }{\|\mathbf b \|_2   } > \frac{  \langle \mathbf x_j , \mathbf a \rangle } { \|\mathbf x_j\|_2  }   \\
	\end{aligned}
	  \right.
\end{equation}

\noindent \textbf{Case 1 } When $\mathbf a=0$, we have $\bm \theta_1^*=\frac{\mathbf y}{\lambda_1}$,
$\mathbf b= \frac{\frac{1}{\lambda_2} - \frac{1}{\lambda_1}}{2} \mathbf y$, $\mathbf x_j^{\bot}=\mathbf x_j$,
and $\mathbf y ^{\bot}=\mathbf y$. Thus, Eq.~\eqref{eq:x:j:upper:bound} can be simplified as:
\begin{equation}
    u_j^+ (\lambda_2)  =\langle \mathbf x_j, \bm \theta_1^* \rangle + \frac{\frac{1}{\lambda_2} - \frac{1}{\lambda_1}}{2}  
 [\|\mathbf x_j^{\bot}\|_2 \|\mathbf y^{\bot}\|_2 + \langle \mathbf x_j^{\bot}, \mathbf y^{\bot} \rangle].
\end{equation}
Since $\|\mathbf x_j^{\bot}\|_2 \|\mathbf y^{\bot}\|_2 + \langle \mathbf x_j^{\bot}, \mathbf y^{\bot} \rangle\ge 0$, 
$u_j^+ (\lambda_2) $ is monotonically decreasing with regard to $\lambda_2$.

\noindent \textbf{Case 2 \& Case 3 }
When $\frac{ \langle \mathbf b, \mathbf a \rangle }{\|\mathbf b \|_2   } = \frac{  \langle \mathbf x_j , \mathbf a \rangle } { \|\mathbf x_j\|_2  }$,
we have 
\begin{equation}
\begin{aligned}
&  \frac{\frac{1}{\lambda_2} - \frac{1}{\lambda_1}}{2}  [ \|\mathbf x_j^{\bot}\|_2 \|\mathbf y^{\bot}\|_2  - \langle \mathbf x_j^{\bot} , \mathbf y ^{\bot} \rangle  ] \\
 & =    \frac{\frac{1}{\lambda_2} - \frac{1}{\lambda_1}}{2}  \left[ \sqrt{( \|\mathbf x_j\|_2^2 - \frac{ \langle \mathbf x_j, \mathbf a \rangle ^2}{  \|\mathbf a\|_2^2} ) 
	                                          ( \|\mathbf y\|_2^2  - \frac{\langle \mathbf y, \mathbf a\rangle ^2}{\|\mathbf a\|_2^2} )}  - [ \langle \mathbf x_j , \mathbf y \rangle
- 
 \frac{\langle  \mathbf a  ,  \mathbf y  \rangle}{\|  \mathbf a  \|_2^2}     \langle \mathbf x_j,  \mathbf a \rangle ]	\right]\\
 & =    \frac{1}{2}  \left[ \sqrt{( \|\mathbf x_j\|_2^2 - \frac{ \|\mathbf x_j\|_2^2  \langle \mathbf b, \mathbf a \rangle ^2}{  \|\mathbf a\|_2^2 \|\mathbf b\|_2^2} ) 
	                                          (\|\mathbf b\|_2^2  - \frac{\langle \mathbf b, \mathbf a\rangle ^2}{\|\mathbf a\|_2^2} )}  - [ \langle \mathbf x_j , \mathbf b - \mathbf a  \rangle
- 
 \frac{\langle  \mathbf a  ,  \mathbf b - \mathbf a  \rangle}{\|  \mathbf a  \|_2^2 }    \langle \mathbf x_j,  \mathbf a \rangle ]	\right]\\
 & =    \frac{1}{2}  \left[  \|\mathbf x_j\|_2 \|\mathbf b\|_2 (1- \frac{\langle \mathbf b, \mathbf a\rangle ^2}{\|\mathbf a\|_2^2 \|\mathbf b\|_2^2})  - [ \langle \mathbf x_j , \mathbf b - \mathbf a  \rangle
- 
 \frac{\langle  \mathbf a  ,  \mathbf b - \mathbf a  \rangle}{\|  \mathbf a  \|_2^2 }    \langle \mathbf x_j,  \mathbf a \rangle ]	\right]\\
 & = \frac{1}{2} [ \|\mathbf x_j\|_2 \|\mathbf b\|_2- \langle \mathbf x_j, \mathbf b \rangle ] + 
\frac{1}{2} [- \frac{ \|\mathbf x_j\|_2  \langle \mathbf b, \mathbf a\rangle ^2}{\|\mathbf a\|_2^2 \|\mathbf b\|_2} +
    \frac{\langle  \mathbf a  ,  \mathbf b   \rangle}{\|  \mathbf a  \|_2^2 }    \langle \mathbf x_j,  \mathbf a \rangle ]\\
 & = \frac{1}{2} [ \|\mathbf x_j\|_2 \|\mathbf b\|_2- \langle \mathbf x_j, \mathbf b \rangle ].
\end{aligned}
\end{equation}
The first equality plugs in the definition of $\mathbf x_j^{\bot}$ and $\mathbf y$ in Eq.~\eqref{eq:x:j:bot} and Eq.~\eqref{eq:y:bot}.
The second equality plugs in  $\frac{ \langle \mathbf b, \mathbf a \rangle }{\|\mathbf b \|_2   } = \frac{  \langle \mathbf x_j , \mathbf a \rangle } { \|\mathbf x_j\|_2  }$,
makes use of Eq.~\eqref{eq:a:b:equation},
and utilizes Eq.~\eqref{eq:b:a:y:relationship:1}.
The last equality further makes use of $\frac{ \langle \mathbf b, \mathbf a \rangle }{\|\mathbf b \|_2   } = \frac{  \langle \mathbf x_j , \mathbf a \rangle } { \|\mathbf x_j\|_2  }$.
The established equality says that $u_j^- (\lambda_2)$ is continuous at 
the $\lambda_2$ that satisfies $\frac{ \langle \mathbf b, \mathbf a \rangle }{\|\mathbf b \|_2   } = \frac{  \langle \mathbf x_j , \mathbf a \rangle } { \|\mathbf x_j\|_2  }$.

It follows from the definition of $\lambda_{2a}$ that if $\lambda_2 \in (\lambda_{2a}, \lambda_1)$ then
$\frac{ \langle \mathbf b, \mathbf a \rangle }{\|\mathbf b \|_2   } > \frac{  \langle \mathbf x_j , \mathbf a \rangle } { \|\mathbf x_j\|_2  }$.
Therefore, according to Eq.~\eqref{eq:minus:x:j:upper:bound}, $u_j^- (\lambda_2)$ is monotonically decreasing with $\lambda_2$ in $(\lambda_{2a}, \lambda_1)$.
Next, we focus on  $\lambda_2$ in the interval $(0, \lambda_{2a}]$.

Denote $\gamma =\frac{1}{\lambda_2} - \frac{1}{\lambda_1}$, and write
$\mathbf b =\frac{\mathbf y}{\lambda_2} - \bm \theta_1^* =
\mathbf a + \gamma \mathbf y $. Thus, $u_j^-(\lambda_2)	 =
	      -\langle \mathbf x_j, \bm \theta_1^* \rangle  +
	      \frac{1}{2} \left[ \|\mathbf x_j\|_2 \|\mathbf b\|_2- \langle \mathbf x_j, \mathbf b \rangle \right] $  can be rewritten as
\begin{equation}
   w(\gamma) = \frac{1}{2}  \left[ \|\mathbf x_j\|_2 \|\mathbf a + \gamma \mathbf y \|_2- \langle \mathbf x_j, \mathbf a + \gamma \mathbf y \rangle \right] 
\end{equation}
The first and second derivatives of $w(\gamma)$ with regard to $\gamma$ can be computed as:
we have
\begin{equation}
   w'(\gamma) = \frac{1}{2}  [ \frac{\|\mathbf x_j\|_2  \langle \mathbf a + \gamma \mathbf y, \mathbf y \rangle}{\|\mathbf a + \gamma \mathbf y \|_2} - \langle \mathbf x_j, \mathbf y \rangle ] 
\end{equation}
\begin{equation}
   w''(\gamma) = \frac{\|\mathbf x_j\|_2  ( \|\mathbf y\|_2^2 \|\mathbf a\|_2^2 - \langle \mathbf a, \mathbf y \rangle^2 )}{2\|\mathbf a + \gamma \mathbf y \|_2^3} \ge 0
\end{equation}
Therefore, we have
\begin{itemize}
	\item  If $ \frac{ \langle \mathbf a, \mathbf y \rangle }{\|\mathbf a\|_2  }  \ge \frac{ \langle \mathbf x_j, \mathbf y \rangle }{\|\mathbf x_j\|_2 }$, i.e., when 
	the angle between $\mathbf y$ and $\mathbf a$ is no larger than the angle between $\mathbf y$ and $\mathbf x_j$, then $w'(\gamma) \ge 0$, and $u_j^- (\lambda_2) $ is monotonically decreasing with regard to $\lambda_2$
	in $(0, \lambda_{2a}]$.
	In this case, the $\lambda_{2a}$ and $\lambda_{2y}$ satisfies $\lambda_{2a} \le \lambda_{2y}$.
	
	\item  If $ \frac{ \langle \mathbf a, \mathbf y \rangle }{\|\mathbf a\|_2  }  < \frac{ \langle \mathbf x_j, \mathbf y \rangle }{\|\mathbf x_j\|_2 }$, 
	let $\gamma_y= \frac{1}{\lambda_{2y}} - \frac{1}{\lambda_1}$. Then, 1) $h'(\gamma_y)=0$, 2) $h'(\gamma_y)<0, \forall 0 < \gamma < \gamma_y$, and $h'(\gamma_y)>0, \forall  \gamma > \gamma_y$.
	Therefore, $u_j^- (\lambda_2) $ is monotonically decreasing with regard to $\lambda_2$ in $(0, \lambda_{2y})$, and monotonically increasing with regard to $\lambda_2$ in $(\lambda_{2y}, \lambda_{2a}]$.
	In this case, the $\lambda_{2a}$ and $\lambda_{2y}$ satisfies $\lambda_{2a} > \lambda_{2y}$.
\end{itemize}

This ends the proof of this theorem. \hfill $\Box$